\documentclass[10pt,twocolumn,letterpaper]{article}

\usepackage{iccv}
\usepackage{times}
\usepackage{epsfig}
\usepackage{graphicx}
\usepackage{amsmath}
\usepackage{amssymb}
\usepackage{graphicx}
\usepackage{amsmath}
\usepackage{amssymb}
\usepackage{booktabs}
\usepackage{color}
\usepackage{multirow}
\usepackage{makecell}
\usepackage{bbm}
\usepackage{bm}
\usepackage{xcolor}         % colors
\usepackage{soul}
\usepackage{newfloat}
\usepackage{listings}
\usepackage{diagbox}
\usepackage{algorithmicx}  
\usepackage[ruled]{algorithm2e} 
\usepackage{colortbl,hhline}
\usepackage{marvosym}

\usepackage[accsupp]{axessibility}

\definecolor{frenchblue}{rgb}{0.0, 0.45, 0.73}
\newcommand{\green}[1]{\textcolor[RGB]{96,187,87}{#1}}
\newcommand{\red}[1]{\textcolor[RGB]{200,78,89}{#1}}
\newcommand{\fn}[1]{\footnotesize{#1}}
\newcommand{\gbf}[1]{\green{\bf{\fn{(#1)}}}}
\newcommand{\rbf}[1]{\red{\bf{\fn{(#1)}}}}
\definecolor{mygray}{gray}{.92}

\usepackage[pagebackref,breaklinks,colorlinks]{hyperref}

\iccvfinalcopy % *** Uncomment this line for the final submission

\def\iccvPaperID{2851} % *** Enter the ICCV Paper ID here

\ificcvfinal\fi

\begin{document}

%%%%%%%%% TITLE
\title{Self-supervised Character-to-Character Distillation for Text Recognition}

\author{Tongkun Guan\textsuperscript{\rm 1}, Wei Shen\textsuperscript{\rm 1\textrm{\Letter}}, Xue Yang\textsuperscript{\rm 1}, Qi Feng\textsuperscript{\rm 2}, Zekun Jiang\textsuperscript{\rm 1}, Xiaokang Yang\textsuperscript{\rm 1}
\\
\textsuperscript{\rm 1} MoE Key Lab of Artificial Intelligence, AI Institute, Shanghai Jiao Tong University\\
\textsuperscript{\rm 2} Department of Automation, Shanghai Jiao Tong University\\
{\tt\small \{gtk0615,wei.shen,yangxue-2019-sjtu,fengqi,zkjiangzekun.cmu,xkyang\}@sjtu.edu.cn}
}

\maketitle
\ificcvfinal\thispagestyle{empty}\fi

%%%%%%%%% ABSTRACT
\begin{abstract}
   When handling complicated text images (e.g., irregular structures, low resolution, heavy occlusion, and uneven illumination), existing supervised text recognition methods are data-hungry. Although these methods employ large-scale synthetic text images to reduce the dependence on annotated real images, the domain gap still limits the recognition performance. Therefore, exploring the robust text feature representations on unlabeled real images by self-supervised learning is a good solution. However, existing self-supervised text recognition methods conduct sequence-to-sequence representation learning by roughly splitting the visual features along the horizontal axis, which limits the flexibility of the augmentations, as large geometric-based augmentations may lead to sequence-to-sequence feature inconsistency. Motivated by this, we propose a novel self-supervised \textbf{C}haracter-to-\textbf{C}haracter \textbf{D}istillation method, CCD, which enables versatile augmentations to facilitate general text representation learning. Specifically, we delineate the character structures of unlabeled real images by designing a self-supervised character segmentation module. Following this, CCD easily enriches the diversity of local characters while keeping their pairwise alignment under flexible augmentations, using the transformation matrix between two augmented views from images.
Experiments demonstrate that CCD achieves state-of-the-art results, with average performance gains of 1.38\% in text recognition, 1.7\% in text segmentation, 0.24 dB (PSNR) and 0.0321 (SSIM) in text super-resolution. Code is available at \url{https://github.com/TongkunGuan/CCD}.
\end{abstract}

%%%%%%%%% BODY TEXT
\section{Introduction}

Recognizing text from images is a fundamental task in computer vision with applications in various scenarios, such as recognizing latex formulas~\cite{Latex}, work-piece serial numbers~\cite{guan2021industrial}, text logos~\cite{Logo}, etc., and contributes significantly to the multi-modal analysis~\cite{multimodal} and text understanding~\cite{TAP,TextVQA}.
However, existing text recognition methods~\cite{wan2020textscanner,ABINET:fang2021read,DAN:wang2020decoupled,ConCLR,MGP,SATRN:lee2020recognizing,S_GTR} are data-hungry, \emph{i.e.}, they require sufficient data with at least text transcriptions for producing accurate character predictions through implicit attention learning~\cite{guan2022glyph}. Even more, some supervised attention methods~\cite{S_GTR,wan2020textscanner,liao2019scene} require extra character-level bounding boxes. These annotations on text images are expensive and laborious. Although they employ text synthesis techniques~\cite{ST:gupta2016synthetic,MJ:jaderberg2014synthetic} to substitute for labour-intensive text annotation tasks, the domain gap between real and synthetic images still limits the performance of text recognition~\cite{DiG}. Therefore, exploring the potential of unlabeled real text images is of great importance as they are readily available.

\begin{figure}[t]
  \centering
  \graphicspath{{./graph/}}
  \includegraphics[width=3.4in]{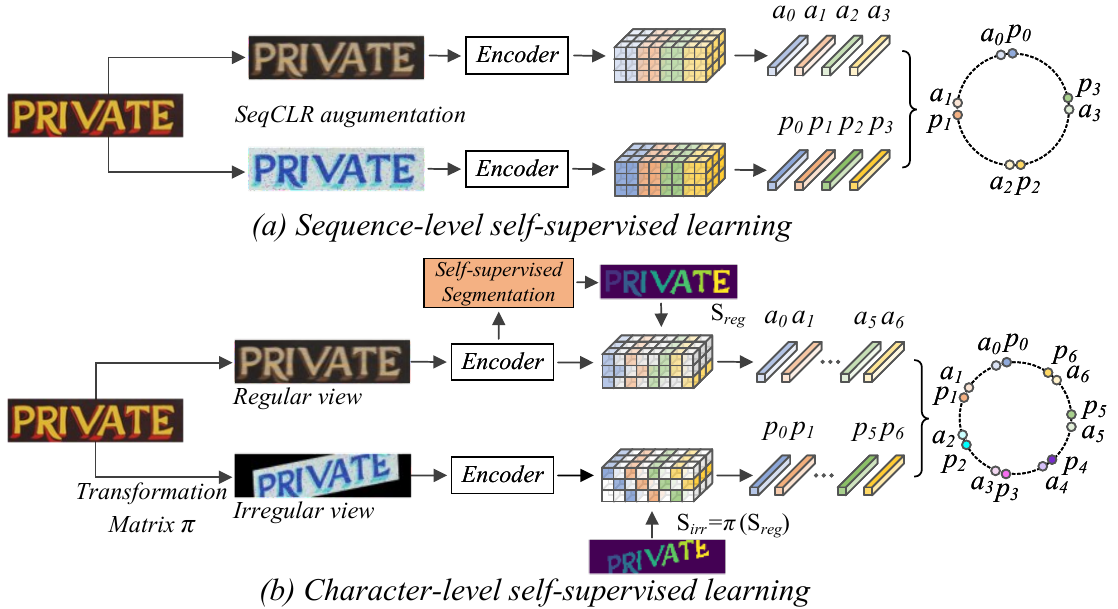}
  \caption{Conceptual illustration of two different self-supervised paradigms. 
  (a) is a sequence-level method that takes feature blocks horizontally-split from a sequence as the basic items for representation learning. (b) is our character-level method that incorporates a self-supervised segmentation head to generate individual character structures ($\mathbf{S}_{reg}$ and $\mathbf{S}_{irr}$), and utilizes the resulting character-level features as the basic items for representation learning.
    }
  \label{Figs.overview}
  \vspace{-1.2em}
\end{figure}

Recently, self-supervised learning methods~\cite{seqCLR,DiG,Persec} have attracted considerable attention, which attempt to leverage the intrinsic qualities of unlabeled real text images to learn proper visual representations, followed by fine-tuning on text-related downstream tasks with less annotated data.
Specifically, SeqCLR~\cite{seqCLR} adopts the SimCLR~\cite{SimCLR} framework to ensure sequence-to-sequence consistency between the two augmented views, in which the sequence is composed of several non-overlapping feature blocks horizontally splitting from the visual features of the text image.
DiG~\cite{DiG} employs both a sequence-level contrastive learning task~\cite{seqCLR} and a masked image modeling task~\cite{MAE} to learn feature representations. 
These methods formulate the learning via sequence-level pretext tasks illustrated in Fig.\ref{Figs.overview} (a). We argue that roughly splitting visual features of text images into a feature sequence along the horizontal axis has two weaknesses: 
1) Inflexible data augmentation strategy, as large geometric transformations may cause inconsistency among the corresponding items in the feature sequence generated from different views. However, versatile data augmentations are demanded for self-supervised learning in many previous works~\cite{SimCLR,asano2019critical,henaff2020data};
2) Neglecting character structures, which confuses networks to cause inter-character mixture, further downgrades the perception of semantic clue information in character-centric text images.
Thus a suitable self-supervised learning paradigm that is tailored for text images with a diversity of word length is in demand.

\begin{figure*}[t]
  \centering
  \graphicspath{{./graph/}}
  \includegraphics[width=6.7in]{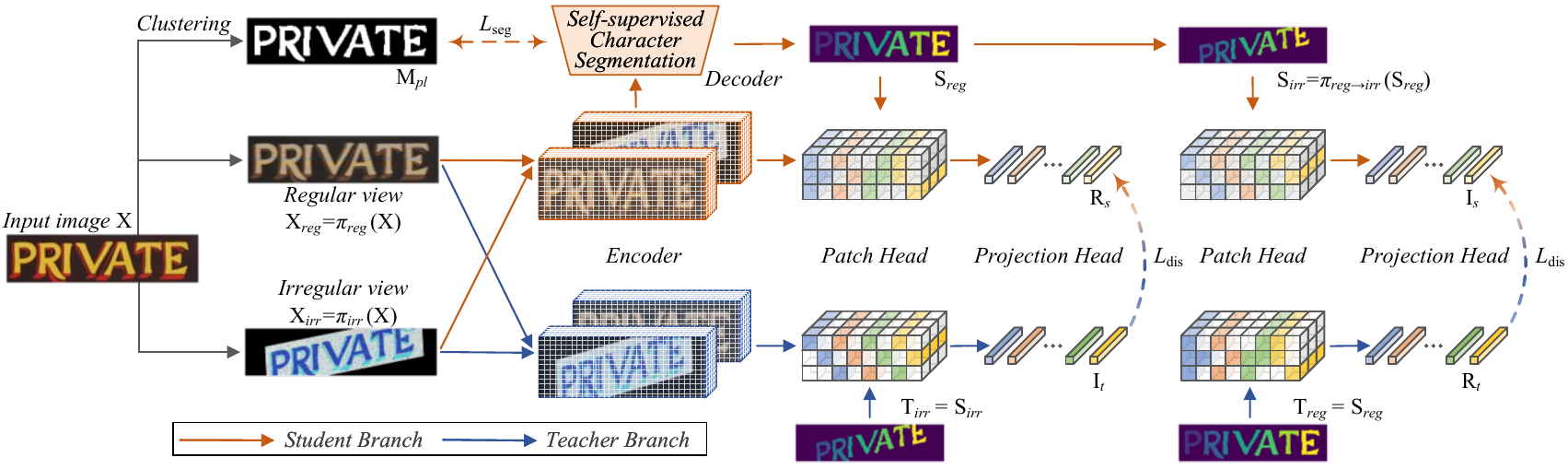}
  \caption{Overview of self-supervised Character-to-Character Distillation (CCD). Both two augmented views ($\mathbf{X}_{reg}$ and $\mathbf{X}_{irr}$) are fed into the student branch and the teacher branch to obtain character features, which are represented by $\mathbf{R}_{s}$ and $\mathbf{I}_{s}$ in the student branch, and $\mathbf{R}_{t}$ and $\mathbf{I}_{t}$ in the teacher branch. We then build character-to-character representation consistency across different views and different branches.
  }
  \label{Figs.Network}
  \vspace{-1.3em}
\end{figure*}

To address this issue, we propose a new self-supervised learning paradigm in character-level, named \textbf{C}haracter-to-\textbf{C}haracter \textbf{D}istillation (CCD), as shown in Fig.\ref{Figs.Network}, which enables feature representation consistency across various augmentations by organizing text images into entities, \emph{i.e.}, each character and background regions. Specifically, two views are first generated from each input image: a regular view with color jitter and an irregular view with additional geometric transformations. Each view is fed into the encoder of the student-teacher branches for extracting features that represent the whole view. Then, character regions from regular view are delineated by a joint self-supervised text segmentation and density-based spatial clustering task, and those from irregular view are generated using the known transformation matrix between two views. In this way, CCD naturally ensures the consistency of corresponding characters across views and branches.
Consequently, by enjoying the pairwise diversity of local characters under flexible augmentations, CCD effectively enhances the robustness and generalization of the learned features, making it more suitable for text-related downstream tasks. In summary, the main contributions are as follows: 
\begin{itemize}
    \setlength{\itemsep}{1pt}
    \setlength{\parsep}{1pt}
    \setlength{\parskip}{1pt}
    \item We propose a novel self-supervised method customized for character-centric text images with a diversity of word length, termed CCD. Different from prior works with sequence-to-sequence pretext tasks, CCD delineates the character structures to establish character-to-character feature representation consistency, enjoying significant augmentation flexibility for extracting general text feature representations.
    \item CCD shows its prominent superiority in self-supervised representation learning, and consistently and significantly outperforms the state-of-the-art DiG~\cite{DiG} by an average of  1.38\% in text recognition, 1.7\% in text segmentation, 0.24 dB (PSNR) and 0.0321 (SSIM) in text image super-resolution task, with the same parameters and latency. 
\end{itemize}

%-------------------------------------------------------------------------
\section{Related Work}

\subsection{Text Recognition}
Given a text image and supervised by its text annotations, text recognition methods aim to predict these characters. Specifically, these methods can be roughly summarized into language-free and language-aware methods. For language-free methods~\cite{CRNN:shi2016end,PIMNet,SAR:li2019show,SATRN:lee2020recognizing,zhong2022sgbanet,wan2020textscanner,liao2019scene,Masktextspotter,Bhunia_2021_ICCV, bhunia2021metahtr,xie2022understanding}, they view text recognition as a character classification task and focus on how to extract robust visual features of characters. For example, some works~\cite{CRNN:shi2016end,PIMNet,SAR:li2019show,SATRN:lee2020recognizing,SCATTER:litman2020scatter,zhong2022sgbanet,ScRN:yang2019symmetry} develop implicit attention mechanisms by computing the similarity of feature patches to extract the important items of visual features, \emph{w.r.t.} the current decoding character. Some works~\cite{wan2020textscanner,liao2019scene,Masktextspotter} employ extra character-level bounding box annotations as the spatial locations of characters to supervise the corresponding attention at the decoding stage, which alleviates the alignment-drifted problem~\cite{robustscanner} and improves attention correctness. For language-aware methods~\cite{ABINET:fang2021read,bhunia2021joint,visionLAN,SRN:yu2020towards}, they execute semantic reasoning on texts with linguistic context to explore the semantic relationships among character, sub-word and word. For example, Fang \emph{et al.}~\cite{ABINET:fang2021read} propose a language model by predicting the masked character in a text with linguistic context. 
Li \emph{et al.}~\cite{visionLAN} mask the visual features of some characters based on character attention to predict the corresponding character categories. 

\subsection{Self-supervised Learning for Text Recognition}
Recently, self-supervised learning~\cite{Dino,SimCLR,IBoT,MAE,MoCo,SimMIM} objectives have gained considerable traction due to their powerful feature representations for transferring downstream tasks. These methods leverage the intrinsic qualities of unlabeled real images to learn general feature representations, by ensuring the feature consistency of various augmented views. 
For example, two popular self-supervised pretext tasks in computer vision are designed to establish representation learning: discriminative task~\cite{seqCLR,MoCo} and generative task~\cite{MAE,SimMIM}. 

Inspired by these methods, some self-supervised text recognition methods~\cite{seqCLR,DiG,Persec,zhang2023relational} are designed for text images with indefinite-length characters, which differs from well-curated images with an atomic input element. 
% object-centric
Specifically, SeqCLR~\cite{seqCLR} is the first to apply a self-supervised method to text recognition, which proposes a sequence-to-sequence contrastive learning framework on text images. It horizontally splits the visual features into a sequence with fixed-length feature blocks, and each item of the feature sequence from two augmented views is aligned. Following SeqCLR with sequence-level representative learning, DiG~\cite{DiG} updates one of the augmented views as a masked view and adds a masked image modeling task.
PerSec~\cite{Persec} performs hierarchical contrastive learning across each element of features for text recognition. 
Our work differs from prior works in that we delineate character structures and propose a character-to-character distillation task to learn more universal text features in the representative space. 

\section{Methodology}

\subsection{Architecture}

\noindent \textbf{Data Augmentation:}
The input image $\mathbf{X}$ undergoes color-based augmentations (\emph{e.g.}, color jitter, color dropout, and grayscale conversion) to create a regular view $\mathbf{X}_{reg}$, and a combination of both color- and geometry-based augmentations (\emph{e.g.}, affine transformation and perspective warping) to generate an irregular view $\mathbf{X}_{irr}$.

\noindent \textbf{Encoder $\mathcal{F(\cdot)}$:} ViT~\cite{ViT} is employed as the encoder of our method CCD due to its prominent superiority in extracting visual features. Specifically, two augmented views ($\mathbf{X}_{reg}$ and $\mathbf{X}_{irr}$) are split into non-overlapping patches with size $4\times4$, and then fed into multi-layer transformer blocks to extract text features. 

\noindent \textbf{Self-supervised Character Segmentation Head $\mathcal{\varPhi(\cdot)}$:} 
We employ an Unet-like network structure to implement pixel-level text segmentation, which assigns each pixel a foreground or background label.
Then, we delineate the character structures from the resulting text segmentation map by a density-based spatial clustering task.

\noindent \textbf{Patch Head $\mathcal{H(\cdot)}$} 
takes the character structure regions and text features as inputs and generates character-level feature representations by a mean-pooling operation.

\noindent \textbf{Projection Head $\mathcal{P(\cdot)}$} consists of a three-layer MLP and a weight-normalized fully connected layer~\cite{Dino}, producing the final character features.

\noindent \textbf{Student \& Teacher Branch}
contain all of the above units, except for self-supervised character segmentation head $\mathcal{\varPhi(\cdot)}$, which is exclusively designed in the student branch to generate the character segmentation results ($\mathbf{S}_{reg}$ in regular view and $\mathbf{S}_{irr}$ in irregular view). 
For simplicity, the segmentation head of the teacher branch is canceled by utilizing the segmentation results ($\mathbf{S}_{reg}$ and $\mathbf{S}_{irr}$) as the corresponding character regions ($\mathbf{T}_{reg}$ in regular view and $\mathbf{T}_{irr}$ in irregular view).
Subsequently, character features ($\mathbf{R}_{s}$ and $\mathbf{I}_{s}$) are generated in the student branch to match the character features ($\mathbf{R}_{t}$ and $\mathbf{I}_{t}$) distribution from the teacher branch. The whole pipeline is illustrated in Fig.\,\ref{Figs.Network} and the detailed network is described in Sec. \ref{section:sec3.2.1}.
Additionally, we also provide a pseudo-code implementation of our character-to-character self-supervised learning method to further illustrate our pipeline as shown in Algo. \ref{algo:CCD}.
\begin{algorithm}[t]
 \caption{Pytorch pseudo-code of CCD.}
 \vspace{-0.5em}
 \label{algo:CCD}
\centering
\definecolor{codeblue}{rgb}{0.25,0.5,0.5}
\definecolor{backcolour}{rgb}{0.95,0.95,0.92}
\definecolor{codegray}{rgb}{0.5,0.5,0.5}
\definecolor{codepurple}{rgb}{0.58,0,0.82}
\definecolor{myMaroon}{rgb}{0.5,0,0}
\definecolor{myOliveGreen}{rgb}{0.2,0.5,0.2}
\definecolor{myDarkBlue}{rgb}{0,0,0.5}
\definecolor{myPurple}{rgb}{0.5,0,0.5}
\definecolor{myDarkGray}{rgb}{0.2,0.2,0.2}
\definecolor{myTeal}{rgb}{0,0.5,0.5}
\definecolor{myDarkRed}{rgb}{0.5,0,0}
\definecolor{myNavyBlue}{rgb}{0,0,0.5}
\definecolor{myDarkGreen}{rgb}{0,0.5,0}
\definecolor{myBrown}{rgb}{0.5,0.2,0}
% \phantomsection
\refstepcounter{table}
\lstset{
  linewidth=0.9\linewidth,      		%列表list宽度
  % backgroundcolor=\color{backcolour},  
  basicstyle=\fontsize{7.2pt}{7.2pt}\ttfamily\bfseries,
  commentstyle=\fontsize{7.2pt}{7.2pt}\color{codeblue},
  % % keywordstyle=\fontsize{7.2pt}{7.2pt}\color{magenta},
  numberstyle=\tiny\color{codegray},
  stringstyle=\color{codepurple},
  % breakatwhitespace=false,         
  breaklines=true,                 
  % captionpos=b,                    
  % keepspaces=true,                 
  % numbers=left,                    
  % numbersep=1pt,    
  xleftmargin=-1.6em, %整体距左侧边线的距离为2em
  % showspaces=false,                
  % showstringspaces=false,
  % showtabs=false,                  
  % tabsize=2,
  literate=%
    {α}{{$\alpha$}}1
    {β}{{$\beta$}}1
    {γ}{{$\gamma$}}1
    {δ}{{$\delta$}}1
    {ε}{{$\epsilon$}}1
    {ζ}{{$\zeta$}}1
    {η}{{$\eta$}}1
    {θ}{{$\theta$}}1
    {ι}{{$\iota$}}1
    {κ}{{$\kappa$}}1
    {λ}{{$\lambda$}}1
    {μ}{{$\mu$}}1
    {ν}{{$\nu$}}1
    {ξ}{{$\xi$}}1
    {ο}{{$\omicron$}}1
    {π}{{$\pi$}}1
    {ρ}{{$\rho$}}1
    {σ}{{$\sigma$}}1
    {τ}{{$\tau$}}1
    {υ}{{$\upsilon$}}1
    {φ}{{$\phi$}}1
    {χ}{{$\chi$}}1
    {ψ}{{$\psi$}}1
    {ω}{{$\omega$}}1
}
\begin{lstlisting}[language=python, mathescape=true, escapeinside={(*}{*)}]
# $\mathcal{G}_s, \mathcal{G}_t$: student and teacher branches
# $\mathcal{F}_s, \mathcal{\varPhi}_s$: encoder and self-supervised character segmentation head in the student branch
$\mathcal{G}_t$.params = $\mathcal{G}_s$.params
for $\mathbf{X}$ in loader: # load a minibatch X
    # augmentation: regular view and irregular view
    $\mathbf{X}_{reg}$, $\mathbf{X}_{irr}$ = augment_reg($\mathbf{X}$), augment_irr($\mathbf{X}$)
    # character regions in the student branch
    $\mathbf{S}_{reg}$ = $\mathcal{\varPhi}_s(\mathcal{F}_s(\mathbf{X}_{reg}))$, $\mathbf{S}_{irr}$ = $\pi_{irr}(\mathbf{S}_{reg})$
    # character regions in the teacher branch
    $\mathbf{T}_{seg}$ = $\mathbf{S}_{reg}$, $\mathbf{T}_{irr}$ = $\mathbf{S}_{irr}$
    # character features in the student branch
    $\mathbf{R}_{s}$, $\mathbf{I}_{s}$ = $\mathcal{G}_s$($\mathbf{X}_{reg}$, $\mathbf{S}_{reg}$), $\mathcal{G}_s$($\mathbf{X}_{irr}$, $\mathbf{S}_{irr}$)
    # character features in the teacher branch
    $\mathbf{R}_{t}$, $\mathbf{I}_{t}$ = $\mathcal{G}_t$($\mathbf{X}_{reg}$, $\mathbf{T}_{seg}$), $\mathcal{G}_t$($\mathbf{X}_{irr}$, $\mathbf{T}_{irr}$)
    loss = ξ($\mathbf{I}_{t}$, $\mathbf{R}_{s}$) + ξ($\mathbf{R}_{t}$, $\mathbf{I}_{s}$)
    loss.backward() # back-propagate
    # student and teacher updates
    update($\mathcal{G}_s$) # AdamW optimizer
    $\mathcal{G}_t$.params = λ*$\mathcal{G}_t$.params + (1 - λ)*$\mathcal{G}_s$.params
\end{lstlisting}
\vspace{-0.5em}
\end{algorithm}

\subsection{Character-level Representation Learning}
In this section, we will illustrate how to obtain character structures and ensure character-to-character consistency for representation learning, which is different from existing sequence-to-sequence self-supervised methods.

\noindent \textbf{1) Self-supervised Character Segmentation.}\label{section:sec3.2.1}
Given an unlabeled text image, our objective is to perform instance-level character segmentation, which identifies all character regions and produces a mask for each of them.
Specifically, to make the task more feasible and reasonable, we divide it into two sub-tasks: a self-supervised text segmentation task and a clustering-based character segmentation task.

For the self-supervised text segmentation task, we first compute the pseudo-labels $\mathbf{M}_{pl}$ of the input images and then use it to train our text segmentation network, which assigns a foreground or background category to each pixel.
To achieve this, we select a simple and effective $K$-means algorithm ~\cite{hartigan1979algorithm} (setting $K$ = 2) to cluster the pixels of each image into a text region (center) and a background
region (surrounding) according to the gray values of pixels. Subsequently, the segmentation network employs the outputs of the 2-nd, 4-th, and 6-th layer of the encoder ViT as $\mathbf{P}_{0}, \mathbf{P}_{1}, \mathbf{P}_{2}$, and the implementation detail is as follows:
{\setlength\abovedisplayskip{1pt}
\setlength\belowdisplayskip{1pt}
\begin{equation}
  \begin{cases}
    \begin{aligned}
      \mathbf{O}_{i} &= \varphi(\mathbf{P}_{i}), i=0,1,2, \\
      \mathbf{O} &= \mathcal{T}([\mathbf{O}_{0}, \mathbf{O}_{1}, \mathbf{O}_{2}]),\\
    \end{aligned}
  \end{cases}
\end{equation}}

\noindent where $\varphi(\cdot)$ denotes two convolutional layers with BatchNorm and ReLU activation functions.
$\mathcal{T}(\cdot)$ refers to two $2 \times$ upsampling operations to restore the resolution of the input image. $[\cdot]$ represents the concatenation operation along the channel axis. 
Finally, the network predictions for text segmentation map $\mathbf{M}_{seg}$ are generated by applying a convolutional layer on $\mathbf{O}$ for binary classification.
A cross-entropy loss $\mathcal{L}_{\rm seg}$ between $\mathbf{M}_{pl}$ and $\mathbf{M}_{seg}$ is employed to optimize our self-supervised text segmentation network. 

For now, let us assume that the text segmentation map $\mathbf{M}_{seg}$ is obtained, the clustering-based character segmentation task aims to get a mask for each of its characters.
One can observe that individual characters retain internal pixel-level connectivity, while the spaces between characters exhibit discontinuities in most natural scene text images. Taking advantage of the observation, we employ a density-based spatial clustering method~\cite{DBSCAN} to
segment the $\mathbf{M}_{seg}$ into several clusters. Specifically, a cluster is formed by aggregating all nearby points connected by density and grouping them together. 
Then, the points within each cluster can be viewed as a character structure. A discussion about hyper-parameters is deferred to ablations and analysis in Sec.\,\ref{Hyper-parameter}.

\begin{figure}[t]
  \centering
  \graphicspath{{./graph/}}
  \includegraphics[width=3.3in]{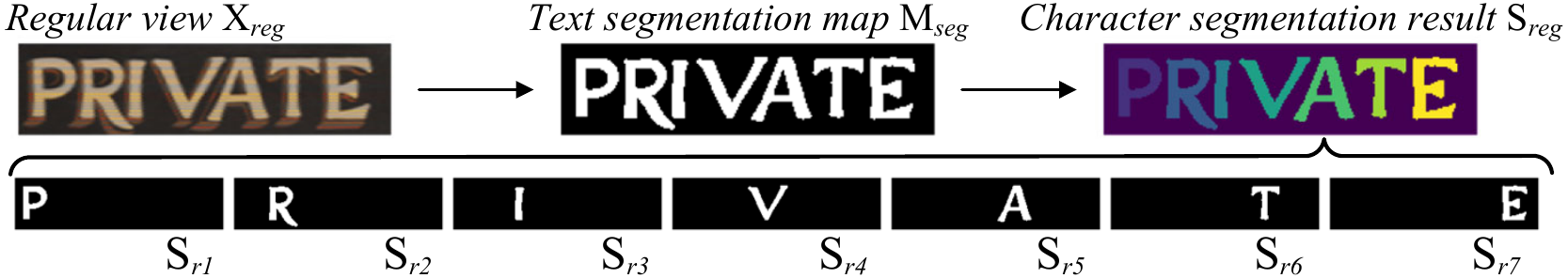}
  \caption{The self-supervised character segmentation pipeline. 
  }
  \label{Figs.DBSCAN}
  \vspace{-1.0em}
\end{figure}

As shown in Fig.\,\ref{Figs.DBSCAN}, the self-supervised character segmentation head exclusively tasks regular views in the student branch as inputs to generate final character segmentation results $\mathbf{S}_{reg} = [\mathbf{s}_{r1}, \mathbf{s}_{r2}, ..., \mathbf{s}_{rl}]$ for simplicity in the experiment. $l$ refers to the number of cluster centers, as well as the word length of the image ideally.

\noindent \textbf{2) Corresponding Regions Alignment.}
An effective data augmentation strategy is crucial to implementing representation learning~\cite{SimCLR,asano2019critical,henaff2020data}. However, in the sequence-level self-supervised text recognition methods, strong geometric transformations will result in the corresponding item inconsistency of feature sequences between different views. 

Motivated by this, we propose an alignment strategy for character regions under flexible augmentations. Specifically,  for the regular view in the student branch, we have obtained character segmentation results $\mathbf{S}_{reg}$ as above. To execute character region alignment, other character segmentation results ($\mathbf{S}_{irr}$, $\mathbf{T}_{reg}$, and $\mathbf{T}_{irr}$) are generated as follows.

To calculate $\mathbf{S}_{irr}$ from the irregular view in the student branch, defined the transformation matrix of augmentation as $\pi$, the problem can be formulated as: \emph{given $\mathbf{X}_{reg} = \pi_{reg}(\mathbf{X}), \mathbf{X}_{irr} = \pi_{irr}(\mathbf{X})$, and $\mathbf{S}_{reg}$, compute $\mathbf{S}_{irr} = \pi_{reg \to irr}(\mathbf{S}_{reg})$}. Consequently, $\mathbf{S}_{irr} = \pi_{irr}(\mathbf{S}_{reg})$ due to $\pi_{reg \to irr} = \pi_{irr}(\pi_{reg}^{-1}) = \pi_{irr}$, where $\pi_{reg}$ is an identity matrix as the regular view undergoes color-based augmentation pipeline. 

For obtaining $\mathbf{T}_{reg}$ from regular views and $\mathbf{T}_{irr}$ from irregular views in the teacher branch respectively, the character segmentation results in the student branch are directly employed, \emph{i.e.}, $\mathbf{T}_{reg} = \mathbf{S}_{reg}, \mathbf{T}_{irr} = \mathbf{S}_{irr}$, due to the same augmentations used in both student and teacher branches.

In this way, our method naturally ensures the alignment of corresponding character regions across different views and branches, which enriches their pairwise diversity. Thus by addressing the challenge of learning feature consistency across diverse augmentations, our method effectively enhances the robustness and generalizability of the learned features, making them more suitable for downstream tasks. 

\noindent \textbf{3) Character-to-character Distillation.}\label{section:sec3.2.3}
With the aforementioned foundation, we can proceed to implement the character-to-character distillation across different views ($\mathbf{X}_{reg}$ and $\mathbf{X}_{irr}$) and branches (student and teacher).

Specifically, we first calculate character feature representations ($\mathbf{R}_{s}$, $\mathbf{I}_{s}$, $\mathbf{R}_{t}$, and $\mathbf{I}_{t}$).
Taking the regular view $\mathbf{X}_{reg}$ in the student branch as an example, we obtain the encoded features $\mathbf{h}_{reg} = \mathcal{F}(\mathbf{X}_{reg})$ and character segmentation results $\mathbf{S}_{reg} = \mathcal{\varPhi}(\mathbf{h}_{reg}) =[\mathbf{s}_{r1}, ..., \mathbf{s}_{rl}]$, the patch head $\mathcal{H}(\cdot)$ further yields the character-level features $\mathbf{V}_{reg} = [\mathbf{v}_{r1}, ..., \mathbf{v}_{rl}]$ by a mean-pooling operation as follows:
 {\setlength\abovedisplayskip{1pt}
\setlength\belowdisplayskip{1pt}
\begin{equation}
    \begin{aligned}
    % \boldsymbol{V}_{R} = [\boldsymbol{V}_{R1}, ..., \boldsymbol{V}_{Rl}] = \boldsymbol{H}(\boldsymbol{h}, \mathbf{S}_{reg}),\\
    \mathbf{v}_{ri} = \frac{1}{\sum_{x,y}s_{ri}^{(x,y)}}\sum_{x,y}s_{ri}^{(x,y)}\mathbf{h}_{reg}^{(x,y)},\\
    \end{aligned}
\end{equation}}
 
\noindent where $i$ represents the channel index, and $(x, y)$ indicates the coordinate point in the map. $\mathbf{V}_{reg}$ is then fed into our projection head $\mathcal{P}(\cdot)$ to get the final character features $\mathbf{R}_{s}=\mathcal{P}(\mathbf{V}_{reg})$ where $\mathbf{R}_{s} \in \mathbb{R}^{l \times n}$. Specifically, $\mathcal{P}(\cdot)$ includes four linear layers, with GELU layer applied to the first two linear layers having a hidden dimension of 2048, and a normalization operation~\cite{Dino} applied to the third linear layer having an output dimension of 256. The last linear layer projects the features into a high-dimension space with n dimensions (n = 65536). 

Following the same principle, we can obtain the remaining character features, \emph{i.e.}, $\mathbf{I}_{s} \in \mathbb{R}^{l \times n}$ from the irregular view in the student branch, $\mathbf{R}_{t}, \mathbf{I}_{t} \in \mathbb{R}^{l \times n}$ from the regular and irregular view in the teacher branch, respectively.

Inspired by DINO~\cite{Dino}, the student is then distilled from the teacher, by optimizing the character-level features ($\mathbf{R}_{s}, \mathbf{I}_{s}$) of the student branch to match those ($\mathbf{R}_{t}, \mathbf{I}_{t}$) of the teacher branch. Assuming $\mathbf{a},\mathbf{b} \in \mathbb{R}^{l \times n}$, let us define:
\begin{small}
  {\setlength\abovedisplayskip{0cm}
  \setlength\belowdisplayskip{0cm}
 \begin{equation}
    \begin{aligned}
    % \rm min (\xi(\mathbf{R}_{t}, \mathbf{I}_{s}) + \xi(\mathbf{I}_{t}, \mathbf{R}_{s})),
    \xi(\mathbf{a},\mathbf{b}) = -\sum_{i=1}^{l}\sum_{j=1}^{n}(\frac{\mathrm{exp}(a_{i}^{j}/\tau)}{\sum_{k=1}^{n}\mathrm{exp}(a_{i}^{k}/\tau)}\,\mathrm{log}\,\frac{\mathrm{exp}(b_{i}^{j}/\tau)}{\sum_{k=1}^{n}\mathrm{exp}(b_{i}^{k}/\tau)})
    \end{aligned}
    \nonumber
\end{equation}}
\end{small}

\noindent The distillation loss is formulated as: $\mathcal{L}_{\rm dis} = \xi(\mathbf{R}_{t}, \mathbf{I}_{s}) + \xi(\mathbf{I}_{t}, \mathbf{R}_{s})$, where $\tau$, a temperature parameter, represents $\tau_{s}$ and $\tau_{t}$ in the student and teacher branches respectively.
Finally, the teacher weights $\theta_{t}$ are updated by applying an exponential moving average (EMA) on the student weights $\theta_{s}$, which is summarized as $\theta_{t} = \lambda \theta_{t} + (1-\lambda) \theta_{s}$.
% \vspace{-0.3em}

\begin{table}[t]
  \caption{The encoder and decoder configurations. ViT-series refers to three variants of ViT (Tiny, Small, and Base).}
  \scalebox{0.73}{
  \begin{tabular}{c|c|clclcl}
\toprule
\multirow{2}{*}{Pre-training Stage } & \multirow{2}{*}{Encoder} & \multicolumn{6}{c}{Encoder Configuration}                                              \\ \cline{3-8} 
                            &                          & \multicolumn{2}{c}{Embed\_dim} & \multicolumn{2}{c}{Depth} & \multicolumn{2}{c}{Heads} \\ \hline
\multirow{3}{*}{CCD}          & ViT-Tiny                 & \multicolumn{2}{c}{192}        & \multicolumn{2}{c}{12}    & \multicolumn{2}{c}{3}     \\ \cline{2-8} 
                            & ViT-Small                & \multicolumn{2}{c}{384}        & \multicolumn{2}{c}{12}    & \multicolumn{2}{c}{6}     \\ \cline{2-8} 
                            & ViT-Base                 & \multicolumn{2}{c}{512}        & \multicolumn{2}{c}{12}    & \multicolumn{2}{c}{8}     \\ \bottomrule
\toprule
\multirow{2}{*}{Fine-tuning Stage}  & \multirow{2}{*}{Encoder} & \multicolumn{6}{c}{Decoder Configuration}                                              \\ \cline{3-8} 
                            &                          & \multicolumn{2}{c}{Embed\_dim} & \multicolumn{2}{c}{Depth} & \multicolumn{2}{c}{Heads} \\ \hline
Text Recognition              & ViT-series               & \multicolumn{2}{c}{512}        & \multicolumn{2}{c}{6}     & \multicolumn{2}{c}{8}     \\ \cline{1-8} 
Text Segmentation             & ViT-Small                & \multicolumn{2}{c}{384}        & \multicolumn{2}{c}{3}     & \multicolumn{2}{c}{2}     \\ \cline{1-8} 
Text Image Super-Resolution   & ViT-Small                & \multicolumn{2}{c}{384}        & \multicolumn{2}{c}{3}     & \multicolumn{2}{c}{2}     \\ \bottomrule
\end{tabular}

  }
  \label{tb:Parameters_Speed}
  \vspace{-1.2em}
\end{table}
\vspace{-0.5em}
\subsection{Down-stream Tasks}
\vspace{-0.5em}
Inherited from the encoder of CCD, the same decoder structures with self-supervised methods~\cite{DiG} for different downstream tasks are employed for fair comparisons. The detailed network configurations are shown in Table \ref{tb:Parameters_Speed}. 

\noindent \textbf{Text Recognition} adds a transformer-based decoder~\cite{SATRN:lee2020recognizing}, which consists of 6 transformer blocks and a linear prediction layer with 96 channels to predict characters.

\noindent \textbf{Text Segmentation} introduces a decoder with 3 transformer blocks and a linear prediction layer. The final dimension is 2 (foreground and background categories).

\noindent \textbf{Text Image Super-Resolution} The same decoder structure with the text segmentation network is employed, except that the final prediction dimension is replaced by 3 to recover the input image with RGB channels.

\section{Experiment}
\subsection{Dataset}

\noindent \textbf{Unlabeled Real Data (URD)} contains 15.77M real-world text images, which are cropped from the large-scale Conceptual Captions Dataset\footnote[1]{https://github.com/google-research-datasets/conceptual-captions} by applying the text bounding box results provided by Microsoft Azure OCR system.

\noindent \textbf{Synthetic Text Data (STD)} consists of two large-scale SynthText~\cite{ST:gupta2016synthetic} (8M) and Synth90k~\cite{MJ:jaderberg2014synthetic} (9M). 

\noindent \textbf{Annotated Real Data (ARD)} is collected from natural scenes and contains 2.78M text images (0.71M in TextOCR~\cite{TextOCR} and 2.07M in Open Image Dataset v5\footnote[2]{https://storage.openvinotoolkit.org/repositories/openvino\_training\_\\extensions/datasets/open\_images\_v5\_text}). 

\noindent \textbf{Scene Text Recognition Benchmarks} include three regular text datasets (\emph{i.e.}, IIIT5K-Words (IIIT) \cite{IIIT5K:mishra2012scene}, ICDAR2013 (IC13) \cite{IC13:karatzas2013icdar}, and Street View Text (SVT) \cite{SVT:wang2011end}) and three irregular text datasets (\emph{i.e.}, ICDAR2015 (IC15) \cite{IC15:karatzas2015icdar}, SVT Perspective (SVTP) \cite{SP:phan2013recognizing}, and CUTE80 (CT) \cite{CT:risnumawan2014robust}).
IIIT, SVT, IC13, IC15, SVTP, and CT benchmarks contain 3000, 647, 1015, 1811, 645, and 288 images, respectively.

\noindent \textbf{Text Segmentation Benchmark} TextSeg~\cite{TextSeg} provides 4024 fine-annotated text images. We crop these images and segmentation maps to construct a text instance segmentation dataset according to its word-level bounding box annotations (training set: 10226, testing set: 3445).

\noindent \textbf{Text Image Super-Resolution Benchmark} 
TextZoom~\cite{TSRN} consists of image pairs with high-resolution and low-resolution text images. Specifically, 17367 image pairs are used for training, while 1619, 1411, and 1343 image pairs are used for evaluation according to difficulty, respectively. 
\begin{table*}[t]
\centering
\caption{Text recognition results compared to other self-supervised text recognizers. * means using an extra 100M images for pre-training.}
\footnotesize
\scalebox{0.95}{
\begin{tabular}{p{23mm}|c|c|p{10mm}p{12mm}p{12mm}p{10mm}p{10mm}p{10mm}|p{14mm}|p{7mm}}
\toprule
Method & Venue & Data & IIIT & SVT & IC13 & IC15 & SVTP & CT &Avg. &Params.\\
\hline
SeqCLR~\cite{seqCLR}  &CVPR'21 &STD & 82.9 &- &87.9 &- &- &- &- & -\\
SimAN~\cite{SimAN} &CVPR'22 &STD &87.5 &- &89.9 &- &- &- &- & -\\
PerSec-ViT$^{*}$~\cite{Persec}&AAAI'22  &STD &88.1 &86.8 &94.2 &73.6 &77.7 &72.7 & 83.77& - \\
\hline
DiG-ViT-Tiny~\cite{DiG}&MM'22           &STD & 95.8 &92.9 &96.4 &84.8 &87.4 &86.1 &91.83 & 20M\\ 
\rowcolor{mygray}
CCD-ViT-Tiny&-           &STD &96.5\gbf{+0.7}  &93.4\gbf{+0.5} &96.3\rbf{-0.1} &85.2\gbf{+0.4} &89.8\gbf{+2.4} &89.2\gbf{+2.9} &92.57\gbf{+0.74} & 20M\\ 
\hline
DiG-ViT-Small~\cite{DiG}&MM'22           &STD & 96.7 &93.4 &97.1 &87.1 &90.1 &88.5 &93.23 & 36M\\
\rowcolor{mygray}
CCD-ViT-Small&-          &STD &96.8\gbf{+0.1}  &94.4\gbf{+1.0} &96.6\rbf{-0.5} &87.3\gbf{+0.2} &91.3\gbf{+1.2} &92.4\gbf{+3.9} &93.59\gbf{+0.37} & 36M\\ 
\hline
DiG-ViT-Base~\cite{DiG}&MM'22           &STD & 96.7 &94.6 &96.9 &87.1 &91.0 &91.3 &93.49 & 52M\\
\rowcolor{mygray}
CCD-ViT-Base &-  &STD &97.2\gbf{+0.5}  &94.4\rbf{-0.2} &97.0\gbf{+0.1} &87.6\gbf{+0.5} &91.8\gbf{+0.8} &93.3\gbf{+2.0} &93.96\gbf{+0.47} & 52M\\ 
\bottomrule
\toprule
DiG-ViT-Tiny~\cite{DiG}&MM'22           &ARD & 96.4 &94.4 &96.2 &87.4 &90.2 &94.1 &93.37 & 20M\\ 
\rowcolor{mygray}
CCD-ViT-Tiny &-          &ARD &97.1\gbf{+0.7}  &96.0\gbf{+1.6} &97.5\gbf{+1.3} &87.5\gbf{+0.1} &91.6\gbf{+1.4} &95.8\gbf{+1.7} &94.18\gbf{+0.81} & 20M\\
\hline
DiG-ViT-Small~\cite{DiG}&MM'22           &ARD &97.7 &96.1 &97.3 &88.6 &91.6 &96.2 &94.69 & 36M\\ 
\rowcolor{mygray}
CCD-ViT-Small  &-         &ARD &98.0\gbf{+0.3}  &96.4\gbf{+0.3} &98.3\gbf{+1.0} &90.3\gbf{+1.7} &92.7\gbf{+1.1} &98.3\gbf{+2.1} &95.57\gbf{+0.88} & 36M\\
\hline
DiG-ViT-Base~\cite{DiG} &MM'22          &ARD & 97.6 &96.5 &97.6 &88.9 &92.9 &96.5 &94.92 & 52M\\ 
\rowcolor{mygray}
CCD-ViT-Base &-          &ARD &98.0\gbf{+0.4} &97.8\gbf{+1.3} &98.3\gbf{+0.7} &91.6\gbf{+2.7}  &96.1\gbf{+3.2} &98.3\gbf{+1.8} &96.30\gbf{+1.38} & 52M\\
\bottomrule
\end{tabular}
}
\label{tb:ssr}
\vspace{-0.5em}
\end{table*}
\begin{table*}[t]
  \centering
  \caption{Comparison results of scene text recognition methods. 
  ``V'' and ``L'' refer to the language-free and language-aware methods, respectively.
  The best results are shown in bold font. 
  ``Avg1" denote the weighted average results of IIIT, SVT, IC13, SVTP and CT by size. ``Avg2" denote the weighted average results of IIIT, SVT, IC15, SVTP and CT by size.
  % Underline values represent the second-best results.
  }
  \scalebox{0.78}{   
    \begin{tabular}{l|c|c|c|p{7mm}<{\centering}p{7mm}<{\centering}p{7mm}<{\centering}p{7mm}<{\centering}p{7mm}<{\centering}p{7mm}<{\centering}|c|c|c|c}
  \toprule
  Methods &Type & Venue &Data & IIIT & SVT & IC13 & IC15 & SVTP & CT & Avg1 & Avg2 &Params. & Time (ms)\\
  \hline
  % SAR~\cite{SAR:li2019show} &\multirow{11}{*}{V} &AAAI2019 &STD &91.5 &84.5 &91.0 &- &76.4  &83.3&- &-\\
  % CA-FCN~\cite{liao2019scene}& &AAAI2019 &STD&91.9 &86.4 &91.5 &- &-&79.9 & &\\
  % DAN~\cite{DAN:wang2020decoupled} &\multirow{8}{*}{V} &AAAI'20 &STD&94.3 &89.2 &93.9 &- &80.0 &84.4 &91.48 &- &-&-\\
  % TextScanner~\cite{wan2020textscanner} & &AAAI2020 &STD&93.9 &90.1 &92.9 &- &84.3 &83.3& &\\
  % SE-ASTER~\cite{2020SEED} & &CVPR2020&STD &93.8 &89.6 &92.8 &- &81.4 &83.6& &\\
  % PlugNet~\cite{mou2020plugnet} & &ECCV2020&STD &94.4 &92.3 &95.0 &- &84.3 &85.0& &\\
  % GA-SPIN~\cite{SPIN} & &AAAI2021&STD &95.2 &90.9 &94.8 &82.8 &83.2 &87.5& &\\
  PIMNet~\cite{PIMNet} &\multirow{8}{*}{V} &MM'21&STD &95.2 &91.2 &93.4 &83.5 &84.3 &84.4 &92.60&89.89&- &28.4\\
  TRBA~\cite{TRBA:baek2019wrong} & &CVPR'21&STD &92.1 &88.9 &93.1 &78.3 &79.5 &78.2&89.74&85.97&50M &27.6\\
  PREN2D~\cite{yan2021primitive} & &CVPR'21&STD &95.6 &94.0 &- &83.0 &87.6 &91.7&-&90.88&- &67.4\\
  % ViTSTR~\cite{VITSTR:atienza2021vision} & &ICDAR2021&STD &88.4 &87.7 &92.4 &78.5 &81.8 &81.3&86M &9.8\\
  Text is Text~\cite{bhunia2021text} & &ICCV'21&STD &92.3 &89.9 &- &- &84.4 &86.3&-&-&- &-\\
  SGBANet~\cite{zhong2022sgbanet} & &ECCV'22&STD &95.4 &89.1&95.1 &- &83.1 &88.2&92.83&-&- &-\\
  CornerTransformer~\cite{xie2022toward} & &ECCV'22&STD &95.9 &94.6 &96.4 &- &91.5 &92.0&95.13&-&86M &294.9\\
  MGP-STR~\cite{MGP}  & &ECCV'22&STD &96.4 &94.7 &- &87.2 &91.0 &90.3&-&92.80&148M &12.3\\
  SIGA~\cite{guan2022glyph}  & &CVPR'23&STD &96.6 &95.1 &96.8 &86.6 &90.5 &93.1&95.58&92.84&113M &56.3\\
  \hline
  SRN~\cite{SRN:yu2020towards}  &\multirow{8}{*}{L} &CVPR'20&STD  &94.8 &91.5 &95.5 &82.7 &85.1  &87.8 &93.07&89.74&49M &26.9\\
  ABINet~\cite{ABINET:fang2021read} & &CVPR'21&STD+WiKi  &96.2 &93.5 &- &86.0 &89.3 &89.2&-&92.02&37M &33.9\\
  JVSR~\cite{bhunia2021joint} & &ICCV'21&STD  &95.2 &92.2 &95.5 &- &85.7 &89.7&93.53&-&44M &26.3\\
  VisionLAN~\cite{visionLAN}  & &ICCV'21&STD   &95.8 &91.7 &95.7 &83.7 &86.0 &88.5&93.80&90.64&33M &-\\
  % S-GTR~\cite{S_GTR}  & &AAAI2022&STD+WiKi &95.1 &93.2 &- &84.1 &86.2 &91.3& &\\
  S-GTR~\cite{S_GTR}  & &AAAI'22&STD+WiKi &95.8 &94.1 &- &84.6 &87.9 &92.3&-&91.50&42M &18.8\\
  ABINet+ConCLR~\cite{ConCLR}  & &AAAI'22&STD+WiKi &96.5 &94.3 &- &85.4 &89.3 &91.3 &-&92.17&- &-\\
  PARSeq~\cite{bautista2022parseq} & &ECCV'22&STD &97.0 &93.6 &96.2 &86.5 &88.9 &92.2 &95.28&92.65&- &-\\
  LevOCR~\cite{LevOCR}  & &ECCV'22&STD &96.6 &92.9 &- &86.4 &88.1 &91.7 &-&92.26&109M &119.0\\
  \hline
  \rowcolor{mygray}
  CCD-ViT-Tiny           & &- &STD &96.5  &93.4 &96.3 &85.2 &89.8 &89.2 &94.96&91.98&20M &43.2 \\
  \rowcolor{mygray}
  CCD-ViT-Small           & &- &STD &96.8  &94.4 &96.6 &87.3 &91.3 &92.4 &95.63&93.11&36M &44.2 \\
  \rowcolor{mygray}
  CCD-ViT-Base           &\multirow{-3}{*}{V} &- &STD &97.2  &94.4 &97.0 &87.6 &91.8 &93.3 &\textbf{96.02}&\textbf{93.48}&52M  &45.0\\
  \hline
  \rowcolor{mygray}
  CCD-ViT-Tiny           & &- &ARD &97.1  &96.0 &97.5 &87.5 &91.6 &95.8 &96.34&93.65&20M  &43.2\\ 
  \rowcolor{mygray}
  CCD-ViT-Small           & &- &ARD &\textbf{98.0} &96.4 &\textbf{98.3} &90.3 &92.7 &\textbf{98.3}  &97.27&95.13&36M  &44.2\\
  \rowcolor{mygray}
  CCD-ViT-Base           &\multirow{-3}{*}{V} &- &ARD &\textbf{98.0} &\textbf{97.8} &\textbf{98.3} &\textbf{91.6}  &\textbf{96.1} &\textbf{98.3} &\textbf{97.83}&\textbf{95.99}&52M  &45.0\\
  \bottomrule
\end{tabular}
    }
\label{tb:results}
\vspace{-1.2em}
\end{table*}

\vspace{-1.2em}
\subsection{Implementation Details}
\noindent \textbf{Self-supervised Pre-Training}
 The pre-training experiments are conducted on URD and STD without labels, at a resolution of $32\times 128$, for fair comparisons. 
 Specifically, we employ ViT-series (\emph{i.e.},  ViT-Tiny,  ViT-Small, and ViT-Base) as the baseline structure of CCD. We train our CCD with the AdamW optimizer~\cite{AdamW}, a cosine learning rate scheduler~\cite{Cosine} with a base learning rate of 5e-4, a cosine weight delay scheduler~\cite{Cosine} from 0.04 to 0.4, batch size with 288, and warm-up for 0.3 epoch in a total of 3 epochs. The temperature $\tau_{s}$ and $\tau_{t}$ are set to 0.1 and 0.04, respectively. The coefficient $\lambda$ follows a cosine scheduler~\cite{Cosine} from 0.996 to 1.

\noindent \textbf{Text Recognition Fine-Tuning}
Our text recognition network is fine-tuned at $32\times 128$ resolution with STD or ARD, and the total training epochs are 10 or 35. The batch size is 384 and the warm-up time is 1 epoch. The same optimizer and learning scheduler are employed.

\noindent \textbf{Text Segmentation Fine-Tuning}
The text segmentation task is fine-tuned at $32\times 128$ resolution with the TextSeg dataset. The batch size is 384, the total number of fine-tuning epochs is 800, and the warm-up time is 50 epochs. The same optimizer and learning scheduler are employed.

\noindent \textbf{Text Image Super-Resolution Fine-Tuning}
The batch size is 384, the total number of fine-tuning epochs is 300, and the warm-up time is 100 epochs. The general Peak Signal-to-Noise Ratio (PSNR) and Similarity Index Measure (SSIM)~\cite{superresolution} are employed to evaluate the quality of super-resolution images. All experiments are implemented on a server with 3 NVIDIA 3090 GPUs in PyTorch.

\subsection{Experiment Results}
\begin{table*}[t]
  \centering
  \caption{The super-resolution evaluation results on the TextZOOM benchmark.}
  % \vspace{-0.4em}
  \scalebox{0.85}{
\begin{tabular}{clccccclccccl}
  \toprule
  &\multirow{2}{*}{Method} & \multirow{2}{*}{Loss} &\multicolumn{4}{c}{SSIM} & & \multicolumn{4}{c}{PSNR}\\ \cline{4-7} \cline{9-12} 
  && &Easy &Medium &Hard &Avg. & &Easy &Medium &Hard &Avg. & \\
  \midrule
  &Bicubic &$\times$ &0.7884 &0.6254 &0.6592 &0.6961 &  &22.35 &18.98 &19.39 &20.35 &\\
  &SRCNN~\cite{SRCNN} &$L_{2}$ &0.8379 &0.6323 &0.6791 &0.7227 &  &23.48 &19.06 &19.34 &20.78 &\\
  &SRResNet~\cite{SRResNet} &$L_{2}$+$L_{tv}$+$L_{p}$ &0.8681 &0.6406 &0.6911 &0.7403 & &24.36 &18.88 &19.29 &21.03 &\\
  &HAN~\cite{HAN} &$L_{2}$ &0.8691 &0.6537 &0.7387 &0.7596 & &23.30 &19.02 &20.16 &20.95 &\\
  &TSRN~\cite{TSRN} &$L_{2}$+$L_{GP}$ &0.8897 &0.6676 &0.7302 &0.7690 & &25.07 &18.86 &19.71 &21.42 &\\
  &TBSRN~\cite{TBSRN} &$L_{POS}$+$L_{CON}$ &0.8729 &0.6455 &0.7452 &0.7603 & &23.46 &19.17 &19.68 &20.91 &\\
  &PCAN~\cite{PCAN} &$L_{2}$+$L_{EG}$ &0.8830 &0.6781 &0.7475 &0.7752 & &24.57 &19.14 &20.26 &21.49 &\\
  \midrule
  &Scratch-ViT-Small &$L_{2}$ &0.8143 &0.6288 &0.6845 &0.7156 & &22.90 &19.65 &20.45 &21.10\\
  &DiG-ViT-Small~\cite{DiG} &$L_{2}$ &0.8613 &0.6561 &0.7215 &0.7522 & &23.98 &19.85 &20.57 &21.60 &\\
  \rowcolor{mygray}
  &CCD-ViT-Small (ours) &$L_{2}$ &0.8822&0.7005&0.7543 &\textbf{0.7843} &&24.40 &20.12 &20.18 &\textbf{21.84} \\
  \bottomrule
  \end{tabular}
  }
  \label{tb:super-resolution}
  \vspace{-1.0em}
  \end{table*}
\begin{table}[t]
  \centering
  \caption{\centering{The text segmentation results on the TextSeg benchmark.}}
  \scalebox{1.0}{
  \begin{tabular}{cccccccc}
  \toprule
  \multirow{2}{*}{Method} & \multicolumn{3}{c}{Scratch-} & \multicolumn{3}{c}{DiG-} & \cellcolor{mygray}CCD-\\
   & \multicolumn{3}{c}{ViT-Small} & \multicolumn{3}{c}{ViT-Small~\cite{DiG}} & \cellcolor{mygray}ViT-Small\\
   \midrule 
   IoU(\%) & \multicolumn{3}{c}{78.1} & \multicolumn{3}{c}{83.1} & \cellcolor{mygray}\textbf{84.8} \\
   \bottomrule
  \end{tabular}
  }
  \label{tb:textsegmentation}
  \vspace{-0.8em}
  \end{table}
\begin{table}[t]
  \centering
  \caption{Ablation study on the text segmentation experiments.}
\begin{tabular}{cccccccc}
  \toprule
  \multirow{2}{*}{Method} & \multicolumn{6}{c}{$K$-means} & \multirow{2}{*}{Self-supervised}\\ 
  \cline{2-7} & \multicolumn{2}{c}{$\bm{\Theta}$} & \multicolumn{2}{c}{$1-\bm{\Theta}$} & \multicolumn{2}{c}{$\mathbf{M}_{pl}$} \\
  \midrule
  IoU(\%) &\multicolumn{2}{c}{39.8} &\multicolumn{2}{c}{40.1} &\multicolumn{2}{c}{\textbf{70.0}} &\textbf{73.6}\\
  \bottomrule
  \end{tabular}
  \label{tb:pseudo label selection}
  \vspace{-1.0em}
\end{table}
\noindent \textbf{Self-supervised text recognition.}
In Table \ref{tb:ssr}, we evaluate the robustness of feature representations for text recognition by comparing the proposed CCD-ViT-series (\emph{i.e.}, Tiny, Small, Base) with previous self-supervised text recognition methods. Our method achieves a new state-of-the-art performance on both regular and irregular benchmarks. 

Specifically, our CCD-ViT-Tiny outperforms the sequence-to-sequence method SeqCLR by 13.6\% and 8.4\% on IIIT and IC13 benchmarks, respectively. Compared to the ``PerSec-ViT*" method that exploits 100M private unlabeled real images for pre-training, CCD-ViT-series is pre-trained on the URD benchmark with 15.77M unlabeled images, and still yields 8.80\%, 9.82\%, and 10.19\% on average accuracy, respectively.

We further conduct a comparison with the previous state-of-the-art self-supervised method, DiG, using the same pre-training data and network parameters. 
Specifically, when fine-tuning with the STD, CCD-ViT-series get better text recognition performances than DiG-ViT-series by 0.74\%, 0.37\%, and 0.47\% on average accuracy, respectively. When fine-tuning with the ARD, CCD-ViT-series consistently and significantly achieves performance gains of 0.81\%, 0.88\%, and 1.38\% on average accuracy compared to DiG-ViT-series. These results demonstrate that our proposed character-level representation learning paradigm is superior to the existing sequence-to-sequence self-learning paradigm, particularly on real-world datasets.
  
\noindent \textbf{Scene text recognition.} 
In Table \ref{tb:results}, we present a comparison between CCD and previous state-of-the-art (SOTA) supervised text recognition methods. Specifically, CCD-ViT-Tiny achieves competitive recognition results (94.96\% vs. 95.58\%) with a minimum parameter of 20M. CCD-ViT-Small gets higher performance and outperforms previous SOTA by 0.27\% on average accuracy while exhibiting a cost-quality trade-off between accuracy and general evaluation metrics (parameter count and latency). CCD-ViT-Base further refreshes the best text recognition results, achieving average gains of 0.44\% and 0.64\% with a smaller model size (52M vs. 113M). When fine-tuning on annotated real data, CCD-ViT-Base attains significantly higher accuracy and establishes new SOTA results, with performance gains by 1.0\%, 2.7\%, 1.5\%, 4.4\%,  4.6\%, and 5.2\% compared to the best results from previous methods on IIIT, SVT, IC13, IC15, SVTP, and CT benchmarks. 
These results underscore the potential of our proposed self-supervised character-level learning as a powerful and flexible method for text recognition.

\noindent \textbf{Text image super-resolution.}
In Table \ref{tb:super-resolution}, CCD is also applied to text image super-resolution task. 
Compared to the Scratch-ViT-Small without the pre-training stage, our CCD-ViT-Small gets better super-resolution results on the SSIM and PSNR metrics. Compared to the self-supervised representation learning method DiG, our method consistently leads performance improvements on the SSIM (0.7843 vs. 0.7522) and PSNR (21.84 vs. 21.60) metrics, while using the same parameters and data for pre-training and fine-tuning.
Notably, our method also achieves higher performance than previous state-of-the-art super-resolution methods, despite employing just three transformer units connected to the ViT-Small structure without additional designs. These experiments show the prominent superiority of our CCD-ViT-Small in improving image quality. 

\noindent \textbf{Text segmentation.}
In the second row of Table \ref{tb:textsegmentation}, our method yields state-of-the-art results and surpasses other methods by a large margin on downstream text segmentation task. Specifically, CCD-ViT-Small surpasses the Scratch-ViT-Small by 6.7\% in terms of Intersection over Union (IoU). Compared with DiG, the first self-supervised learning method to evaluate text segmentation performance on text images, our method shows its prominent superiority, with a performance improvement of 1.7\% IoU. These experimental results demonstrate our self-supervised method can learn more generalized text feature representations.
\vspace{-0.5em}
\section{Ablations and analysis}
\begin{algorithm}[t]
\small
     \caption{The text pseudo-label $\mathbf{M}_{pl}$ selection.}
     \label{training_BoxDQN}
     Denote a clustering result $\bm{\Theta} \in
\{0, 1\}^{H\times W}$.\\
    Define $\Theta_{i,j}$ is the pixel value of 
    row i, column j.\\
    Calculate the sum of pixel values for each side:\\
     $L$ = $\sum_{i=1}^{H}\Theta_{i,1}$;\,\,\,
     $R$ = $\sum_{i=1}^{H}\Theta_{i,W}$;\\
     $T$ = $\sum_{j=1}^{W}\Theta_{1,j}$;\,\,
     $B$ = $\sum_{j=1}^{W}\Theta_{H,j}$; \\
     Get condition $\varGamma$:\\
     $\varGamma = \mathbbm{1}_{[T \geqslant \frac{W}{2}]} + \mathbbm{1}_{[B \geqslant \frac{W}{2}]} + \mathbbm{1}_{[L \geqslant \frac{H}{2}]} + \mathbbm{1}_{[R \geqslant \frac{H}{2}]},$\\
     \eIf{$\varGamma >= 3$:}
             {
                 $\mathbf{M}_{pl}=1-\bm{\Theta}$
             }
             {
                 $\mathbf{M}_{pl}=\bm{\Theta}$
             }
    \vspace{-0.2em}
\end{algorithm}
\begin{table*}[t]
  \centering
  \caption{Feature representation evaluation of CCD on scene text recognition benchmarks.}
    \begin{tabular}{l|ccccccccccc|c}
    \toprule
    Method & IIIT & SVT & IC13 & IC15 & SVTP & CUTE & COCO & CTW & TT &HOST &WOST &Avg.\\
    \midrule
    Gen-ViT-Small &86.6 &82.1 &88.7 &72.9 &74.4 & 72.2 &48.5&64.1 &63.3&33.8&56.5&59.3\\
    Dis-ViT-Small &92.6 &90.4 &93.4 &81.2 &81.7 & 84.0 &60.0&72.8 &73.1&33.3&56.1&67.0\\
    DiG-ViT-Small &94.2 &93.0 &95.3 &84.3 &86.1 &87.5&63.4&77.9 &75.8&41.7&64.0&\textbf{71.1}\\
    \rowcolor{mygray}
    CCD-ViT-Small &93.5 &89.6 &92.8 &82.7 &85.1 &83.0 &60.4 &73.3&73.4&47.6&66.5&69.9  \\
    \bottomrule
    \end{tabular}
  \label{tb:feature evaluation}
  \vspace{-0.5em}
  \end{table*}
\begin{table*}[t]
  \centering
  \caption{Comparison results when training with different data ratios.}
  \scalebox{0.88}{
    \begin{tabular}{ll|ccccccccccc|c}
    \toprule
    Label Fraction& Method & IIIT & SVT & IC13 & IC15 & SVTP & CUTE & COCO & CTW & TT &HOST &WOST &Avg.\\
    \midrule
    \multirow{2}{*}{1\%(27.8K)} &DiG-ViT-Small &88.4 &86.2 &89.9 &79.0 &76.6 &77.8&54.8&67.9&67.2&33.2&53.3&62.9\\
    & \cellcolor{mygray}CCD-ViT-Small &\cellcolor{mygray}89.3 &\cellcolor{mygray}86.5 &\cellcolor{mygray}88.8 &\cellcolor{mygray}76.5 &\cellcolor{mygray}80.1 &\cellcolor{mygray}74.7 &\cellcolor{mygray}54.9 &\cellcolor{mygray}65.5 &\cellcolor{mygray}67.8 &\cellcolor{mygray}38.4 &\cellcolor{mygray}55.9 &\cellcolor{mygray}\textbf{63.7}  \\
    \midrule
    \multirow{2}{*}{10\%(278K)}&DiG-ViT-Small &95.3 &94.4 &95.9 &85.3 &87.9 &91.7&67.1&80.5 &81.1&42.1&64.0&73.5\\
    &\cellcolor{mygray}CCD-ViT-Small &\cellcolor{mygray}95.9 &\cellcolor{mygray}94.1 &\cellcolor{mygray}96.6 &\cellcolor{mygray}87.1 &\cellcolor{mygray}89.9 &\cellcolor{mygray}94.1 &\cellcolor{mygray}69.2 &\cellcolor{mygray}81.6 &\cellcolor{mygray}84.3 &\cellcolor{mygray}63.4 &\cellcolor{mygray}76.2 &\cellcolor{mygray}\textbf{78.2}  \\
    \midrule
    \multirow{2}{*}{100\%(2.78M)}&DiG-ViT-Small &97.7 &96.1 &97.3 &88.6 &91.6 &96.2&75.0&86.3 &88.9&56.0&75.7&80.7\\
    &\cellcolor{mygray}CCD-ViT-Small &\cellcolor{mygray}98.0 &\cellcolor{mygray}96.4 &\cellcolor{mygray}98.3 &\cellcolor{mygray}90.3 &\cellcolor{mygray}92.7 &\cellcolor{mygray}98.3 &\cellcolor{mygray}76.7 &\cellcolor{mygray}86.5 &\cellcolor{mygray}91.3 &\cellcolor{mygray}77.3 &\cellcolor{mygray}86.0 &\cellcolor{mygray}\textbf{84.9}  \\
    \bottomrule
    \end{tabular}
    }
  \label{tb:Fune-tuning evaluation}
  \vspace{-1.0em}
  \end{table*}

\noindent \textbf{Selection of text pseudo-labels.}
The clustering result provided by the $K$-means algorithm may correspond to a text region or a background region. Thus, as illustrated in Algo.\ref{training_BoxDQN}, we make a minor adaptation to select appropriate text regions from the clustering results by leveraging the observation that text regions are typically located in the center of most scene text instance images, with the four sides of the image being mainly background regions. Compared with randomly selected clustering results ($\bm{\Theta}$ or $1-\bm{\Theta}$), this adaptation results $\mathbf{M}_{pl}$ leads to a 29.9\% IoU performance gain (70.0\% vs. 40.1\%), as shown in Table \ref{tb:pseudo label selection}.

\noindent \textbf{Effectiveness of self-supervised text segmentation.}
In the first row of Table \ref{tb:pseudo label selection}, our self-supervised text segmentation network achieves a 3.6\% IoU improvement (73.6\% vs. 70.0\%) by utilizing the text regions clustered by $K$-means as pseudo-labels. Additionally, when $K$-means is employed for character-to-character representation learning, the average accuracy understandably decreases by 0.24\% in row B of Table \ref{tb:ablation_study}.
This can be attributed to two reasons:
\textbf{1)} Neural network has the ability to learn generality from massive training data, which alleviates the noise in the pseudo-labels introduced by $K$-means. \textbf{2)} The underlying morphological representations of glyphs that we need are relatively invariant to slight structural changes, \emph{e.g.}, thicker or thinner, which reduces the dependence on pixel-level high-precision segmentation with expensive costs. 

\noindent \textbf{Feature representation evaluation.}
Referring to DiG, we freeze the encoder and train the decoder with ARD. As shown in Table \ref{tb:feature evaluation}, our result is 1.2\% lower than DiG (71.1\% \emph{vs} 69.9\%). However, this result is not closely related to the final STR results. As evidenced in Table 2 and 3 of DiG, although the discriminative task ``Dis-$\ast$" outperforms the generative task ``Gen-$\ast$" by 7.7\%, it only shows comparable performance to ``Gen-$\ast$" in the final STR results. Besides, the reason behind DiG's superior result is attributed to their utilization of both ``Gen-$\ast$" and ``Dis-$\ast$", which brings 4.1\% gains. Therefore, when we fairly compare to the same type of pretext task ``Dis-$\ast$",  CCD gets a gain of 2.9\% (69.9\% \emph{vs} 67.0\%). Even compared to DiG, Ours improve by 5.9\% (47.6\% \emph{vs} 41.7\%) and 2.5\% (66.5\% \emph{vs} 64.0\%) on the challenging occluded datasets HOST and WOST, respectively.

\noindent \textbf{Fine-tuning with different data ratios.}
To demonstrate the effectiveness of our proposed CCD, we further fine-tuned our method with 1\%, 10\% and 100\% of ARD. As shown in Table \ref{tb:Fune-tuning evaluation}, CCD outperforms DiG by 0.8\%, 4.7\% and 4.2\%, respectively.

\noindent \textbf{Effectiveness of distillation strategy.} 
The distillation strategy serves as a fundamental component in self-supervised representation learning.
To demonstrate the effectiveness of the distillation strategy, we consider CCD without distillation as the baseline model. Specifically, we add the result as shown in row A of Table \ref{tb:ablation_study}. Compared to Baseline, the total improvement is 4.12\%.

\noindent \textbf{Effectiveness of augmentation strategy.} 
In Table \ref{tb:ablation_study}, we compare two augmentation strategies used to establish character-to-character representation consistency across regular and irregular views. 1) ``R2R'': both the regular and irregular views adopt color-based augmentations; 2) ``CCD'': the default setting, as regular views adopt color-based augmentations, and irregular views adopt a combination of both color- and geometry-based augmentations.

Compared with ``R2R'' method, the latter achieves 0.49\% improvements, which demonstrates large geometric augmentations are still suitable for text images with a diversity of word length, and encourage CCD to reach a data-efficient self-supervised learning regime.

\noindent \textbf{More comparison results.}
1) We first compared with the language-aware PARSeq (0.5\% gains) in Table \ref{tb:results} when training with STD. 
Besides, we also conduct the comparative experiment using real data provided by PARSeq, resulting in 0.35\% gains (+.0, +.4, +.3, +.2, +.5, +.6, +.7, +1.0, respectively).
2) Compared to using the cropped patches, CCD achieves 0.7\%, 1.1\%, 0.0\%, 0.7\%, 4.3\% and 2.0\% gains on six standard benchmarks, respectively.

\noindent \textbf{Further discussion on data-hungry nature of models.}
Generally, existing methods achieve significant improvements when training on ARD compared to STD. However, the amount of ARD (only 2.78M) pales in comparison to massive and readily available unlabeled data. Therefore, CCD provides an effective solution by leveraging the intrinsic qualities of unlabeled data (15.77M) to extract robust feature representations, which can generalize well on multiple text-related downstream tasks.

\begin{table}[t]
\centering
\caption{\centering{Ablation results on scene text recognition benchmarks.}}
\scalebox{0.78}{
\begin{tabular}{l|l|p{5mm}<{\centering}p{5mm}<{\centering}p{5mm}<{\centering}p{5mm}<{\centering}p{6mm}<{\centering}p{5mm}<{\centering}|l}
\toprule
Index & Method & IIIT & SVT & IC13 & IC15 & SVTP & CT & Avg. \\ 
\midrule
A & Baseline &95.0 &92.9 &94.9 &85.2 &86.7 &88.9 &91.45\\
\midrule
B & $K$-means &97.6 &96.2 &98.0 &90.5 &92.3 &97.4 &95.33\\
\midrule
C & R2R &97.8 &96.4 &97.3 &89.5 &92.7 &96.4 &95.08 \\
\midrule
\rowcolor{mygray}
E & CCD  &98.0 &96.4 &98.3 &90.3 &92.7 &98.3 &\textbf{95.57} \\
\bottomrule
\end{tabular}
}
\label{tb:ablation_study}
\end{table}

\begin{table}[t]
  \centering
  \caption{\centering{Quantification results for different types of clusters.}}
  \scalebox{0.83}{
  \begin{tabular}{cccccc}
  \toprule
   Types &background &string &semi-character & character\\
  \midrule
  Number &1583 &991 &1802 &\textbf{14665} \\
  \bottomrule
  \end{tabular}
  }
  \label{tb:clusters}
  \end{table}
  
\begin{figure}[t]
\centering
\graphicspath{{./graph/}}
\includegraphics[width=3.4in]{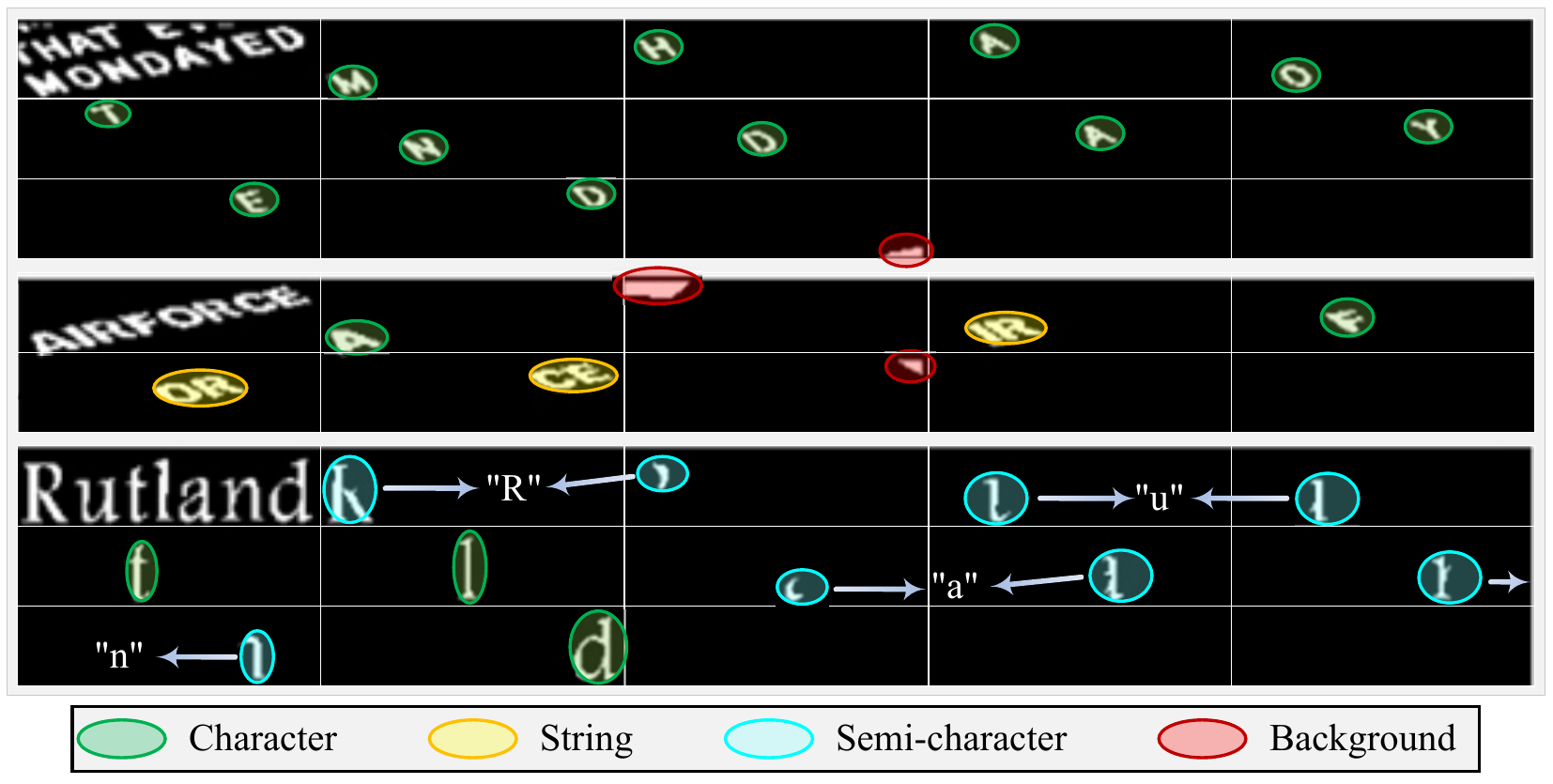}
\caption{Representative clustering visualization results.}
\label{Figs.failure_examples}
\end{figure}

\noindent \textbf{Further discussion on density-based spatial clustering.}
The density-based spatial clustering method may encounter difficulties in separating interconnected characters into isolated clusters and has a proneness to cluster an individual character into multiple clusters due to the different densities of connected regions as shown in Fig.\,\ref{Figs.failure_examples}. Such failures are typically observed in images with tightly interconnected strokes and can also arise due to inaccuracies in text mask predictions generated by our self-supervised text segmentation network. To provide a detailed statistical illustration, we count the number of different types of clusters evaluated on the TextSeg dataset in Table \ref{tb:clusters}. The result was obtained through 7 hours of meticulous manual counting. Clearly, these failures are rare while most clusters still belong to complete characters.

\begin{figure}[t]
  \centering
  \graphicspath{{./graph/}}
  \includegraphics[width=3.4in]{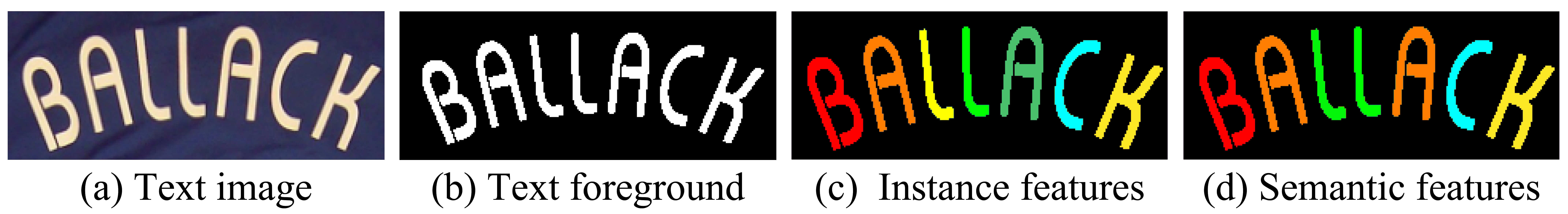}
  \caption{\centering{Three different types of text feature representations.}}
  \label{Figs.three_learning_capability}
\end{figure}

For further discussion, self-supervised learning objective is to obtain general text feature representations, \emph{e.g.,} text foreground features (b) or even character instance features (c) in Fig.\,\ref{Figs.three_learning_capability}, while the latter takes a step forward than the former, towards semantic features (d) extracted by supervised learning regimes. 
As presented above, CCD can obtain more complete characters, enjoying approximately 77\% pairwise character alignment under flexible augmentations, towards learning character instance feature representations (c). 
Even in situations where it struggles to cluster characters due to the issues mentioned above, representation consistency in string or semi-character (both belonging to foreground compared with an item of sequence) can still effectively facilitate the learning of text foreground features (b). 
Overall, our method can at least degenerate into the same representation consistency learning with sequence-level self-supervised methods in worse-case scenarios, which demonstrates the feasibility and effectiveness of our CCD. 
More visualization examples about $\mathbf{M}_{pl}$, $\mathbf{M}_{seg}$, and $\mathbf{S}_{reg}$ are shown in Supplementary Material.

\begin{figure}[t]
  \centering
  \graphicspath{{./graph/}}
  \includegraphics[width=3.2in]{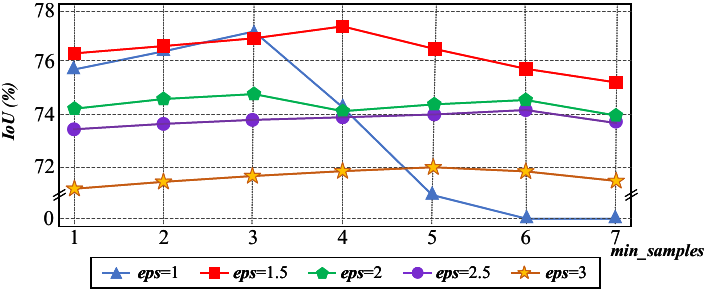}
  \caption{Ablation study of the hyper-parameters.}
  \label{Figs.parameter}
\end{figure}

\noindent \textbf{Hyper-parameters.} \label{Hyper-parameter}
The density-based spatial clustering method is sensitive to the parameters, namely, $\emph{eps}$ and $\emph{min\_samples}$. The $\emph{eps}$ controls the granularity of the clustering, while the $\emph{min\_samples}$ sets a threshold for the minimum number of points required for a cluster to be formed.
To achieve optimal clustering performance, we conduct a grid search over a range of parameter values and select the best combination based on the IoU metric, as illustrated in Fig.\,\ref{Figs.parameter}. Specifically, we first apply this clustering method to the text masks in the training set of Textseg to obtain character clustering results, and then calculate the IoU between these results and the annotated character structures.

\section{Conclusion}
In this paper, we propose a novel self-supervised text recognition method in character-level, termed CCD, which ensures character-to-character representation consistency under flexible augmentations by keeping their pairwise alignment of character regions.
Different from existing sequence-to-sequence self-supervised learning models, CCD takes the delineated character structures as basic items for representation learning, and proposes an effective augmentation strategy to enrich the diversity of local character regions. Eventually, CCD shows significant improvement in the robustness and generalizability of the extracted feature representations and refreshes state-of-the-art performance on three text-related tasks. 

\noindent \textbf{Acknowledgements} This work was supported by NSFC 62176159, Natural Science Foundation of Shanghai 21ZR1432200, Shanghai Municipal Science and Technology Major Project 2021SHZDZX0102 and the Fundamental Research Funds for the Central Universities.

\newpage
{\small
\bibliographystyle{ieee_fullname}
\bibliography{egbib}

\begin{thebibliography}{10}\itemsep=-1pt

\bibitem{seqCLR}
Aviad Aberdam, Ron Litman, Shahar Tsiper, Oron Anschel, Ron Slossberg, Shai
  Mazor, R. Manmatha, and Pietro Perona.
\newblock Sequence-to-sequence contrastive learning for text recognition.
\newblock In {\em CVPR}, pages 15302--15312, June 2021.

\bibitem{asano2019critical}
Yuki~M Asano, Christian Rupprecht, and Andrea Vedaldi.
\newblock A critical analysis of self-supervision, or what we can learn from a
  single image.
\newblock {\em arXiv preprint arXiv:1904.13132}, 2019.

\bibitem{TRBA:baek2019wrong}
Jeonghun Baek, Geewook Kim, Junyeop Lee, Sungrae Park, Dongyoon Han, Sangdoo
  Yun, Seong~Joon Oh, and Hwalsuk Lee.
\newblock What is wrong with scene text recognition model comparisons? dataset
  and model analysis.
\newblock In {\em ICCV}, pages 4715--4723, 2019.

\bibitem{bautista2022parseq}
Darwin Bautista and Rowel Atienza.
\newblock Scene text recognition with permuted autoregressive sequence models.
\newblock In {\em ECCV}, volume 13688, pages 178--196, 2022.

\bibitem{Bhunia_2021_ICCV}
Ayan~Kumar Bhunia, Pinaki~Nath Chowdhury, Aneeshan Sain, and Yi-Zhe Song.
\newblock Towards the unseen: Iterative text recognition by distilling from
  errors.
\newblock In {\em ICCV}, pages 14950--14959, October 2021.

\bibitem{bhunia2021metahtr}
Ayan~Kumar Bhunia, Shuvozit Ghose, Amandeep Kumar, Pinaki~Nath Chowdhury,
  Aneeshan Sain, and Yi-Zhe Song.
\newblock Metahtr: Towards writer-adaptive handwritten text recognition.
\newblock In {\em CVPR}, pages 15830--15839, 2021.

\bibitem{bhunia2021text}
Ayan~Kumar Bhunia, Aneeshan Sain, Pinaki~Nath Chowdhury, and Yi-Zhe Song.
\newblock Text is text, no matter what: Unifying text recognition using
  knowledge distillation.
\newblock In {\em ICCV}, pages 983--992, 2021.

\bibitem{bhunia2021joint}
Ayan~Kumar Bhunia, Aneeshan Sain, Amandeep Kumar, Shuvozit Ghose, Pinaki~Nath
  Chowdhury, and Yi-Zhe Song.
\newblock Joint visual semantic reasoning: Multi-stage decoder for text
  recognition.
\newblock In {\em ICCV}, pages 14940--14949, 2021.

\bibitem{Dino}
Mathilde Caron, Hugo Touvron, Ishan Misra, Herv{\'{e}} J{\'{e}}gou, Julien
  Mairal, Piotr Bojanowski, and Armand Joulin.
\newblock Emerging properties in self-supervised vision transformers.
\newblock In {\em ICCV}, pages 9630--9640, 2021.

\bibitem{TBSRN}
Jingye Chen, Bin Li, and Xiangyang Xue.
\newblock Scene text telescope: Text-focused scene image super-resolution.
\newblock In {\em CVPR}, pages 12026--12035, 2021.

\bibitem{SimCLR}
Ting Chen, Simon Kornblith, Mohammad Norouzi, and Geoffrey~E. Hinton.
\newblock A simple framework for contrastive learning of visual
  representations.
\newblock In {\em ICML}, volume 119, pages 1597--1607, 2020.

\bibitem{LevOCR}
Da Cheng, Peng Wang, and Cong Yao.
\newblock Levenshtein ocr.
\newblock In {\em ECCV}, volume 13688, pages 322--338, 2022.

\bibitem{SRCNN}
Chao Dong, Chen~Change Loy, Kaiming He, and Xiaoou Tang.
\newblock Image super-resolution using deep convolutional networks.
\newblock {\em IEEE TPAMI}, 38(2):295--307, 2016.

\bibitem{ViT}
Alexey Dosovitskiy, Lucas Beyer, Alexander Kolesnikov, Dirk Weissenborn,
  Xiaohua Zhai, Thomas Unterthiner, Mostafa Dehghani, Matthias Minderer, Georg
  Heigold, Sylvain Gelly, Jakob Uszkoreit, and Neil Houlsby.
\newblock An image is worth 16x16 words: Transformers for image recognition at
  scale.
\newblock In {\em ICLR}, 2021.

\bibitem{DBSCAN}
Martin Ester, Hans-Peter Kriegel, J{\"o}rg Sander, Xiaowei Xu, et~al.
\newblock A density-based algorithm for discovering clusters in large spatial
  databases with noise.
\newblock In {\em KDD}, volume~96, pages 226--231, 1996.

\bibitem{ABINET:fang2021read}
Shancheng Fang, Hongtao Xie, Yuxin Wang, Zhendong Mao, and Yongdong Zhang.
\newblock Read like humans: Autonomous, bidirectional and iterative language
  modeling for scene text recognition.
\newblock In {\em CVPR}, pages 7098--7107, 2021.

\bibitem{Cosine}
Jean{-}Bastien Grill, Florian Strub, Florent Altch{\'{e}}, Corentin Tallec,
  Pierre~H. Richemond, Elena Buchatskaya, Carl Doersch, Bernardo~{\'{A}}vila
  Pires, Zhaohan Guo, Mohammad~Gheshlaghi Azar, Bilal Piot, Koray Kavukcuoglu,
  R{\'{e}}mi Munos, and Michal Valko.
\newblock Bootstrap your own latent - {A} new approach to self-supervised
  learning.
\newblock In {\em NeurIPS}, 2020.

\bibitem{guan2021industrial}
Tongkun Guan, Chaochen Gu, Changsheng Lu, Jingzheng Tu, Qi Feng, Kaijie Wu, and
  Xinping Guan.
\newblock Industrial scene text detection with refined feature-attentive
  network.
\newblock {\em IEEE TCSVT}, 32(9):6073--6085, 2022.

\bibitem{guan2022glyph}
Tongkun Guan, Chaochen Gu, Jingzheng Tu, Xue Yang, Qi Feng, Yudi Zhao, and Wei
  Shen.
\newblock Self-supervised implicit glyph attention for text recognition.
\newblock In {\em CVPR}, pages 15285--15294, 2023.

\bibitem{ST:gupta2016synthetic}
Ankush Gupta, Andrea Vedaldi, and Andrew Zisserman.
\newblock Synthetic data for text localisation in natural images.
\newblock In {\em CVPR}, pages 2315--2324, 2016.

\bibitem{hartigan1979algorithm}
John~A Hartigan and Manchek~A Wong.
\newblock Algorithm as 136: A k-means clustering algorithm.
\newblock {\em Journal of the royal statistical society. series c (applied
  statistics)}, 28(1):100--108, 1979.

\bibitem{MAE}
Kaiming He, Xinlei Chen, Saining Xie, Yanghao Li, Piotr Doll{\'{a}}r, and
  Ross~B. Girshick.
\newblock Masked autoencoders are scalable vision learners.
\newblock In {\em CVPR}, pages 15979--15988. {IEEE}, 2022.

\bibitem{MoCo}
Kaiming He, Haoqi Fan, Yuxin Wu, Saining Xie, and Ross~B. Girshick.
\newblock Momentum contrast for unsupervised visual representation learning.
\newblock In {\em CVPR}, pages 9726--9735, 2020.

\bibitem{S_GTR}
Yue He, Chen Chen, Jing Zhang, Juhua Liu, Fengxiang He, Chaoyue Wang, and Bo
  Du.
\newblock Visual semantics allow for textual reasoning better in scene text
  recognition.
\newblock In {\em AAAI}, volume~36, pages 888--896, 2022.

\bibitem{henaff2020data}
Olivier Henaff.
\newblock Data-efficient image recognition with contrastive predictive coding.
\newblock In {\em ICML}, pages 4182--4192, 2020.

\bibitem{MJ:jaderberg2014synthetic}
Max Jaderberg, Karen Simonyan, Andrea Vedaldi, and Andrew Zisserman.
\newblock Synthetic data and artificial neural networks for natural scene text
  recognition.
\newblock {\em arXiv preprint arXiv:1406.2227}, 2014.

\bibitem{IC15:karatzas2015icdar}
Dimosthenis Karatzas, Lluis Gomez-Bigorda, Anguelos Nicolaou, Suman Ghosh,
  Andrew Bagdanov, Masakazu Iwamura, Jiri Matas, Lukas Neumann,
  Vijay~Ramaseshan Chandrasekhar, Shijian Lu, et~al.
\newblock Icdar 2015 competition on robust reading.
\newblock In {\em ICDAR}, pages 1156--1160. IEEE, 2015.

\bibitem{IC13:karatzas2013icdar}
Dimosthenis Karatzas, Faisal Shafait, Seiichi Uchida, Masakazu Iwamura,
  Lluis~Gomez i Bigorda, Sergi~Robles Mestre, Joan Mas, David~Fernandez Mota,
  Jon~Almazan Almazan, and Lluis~Pere De~Las~Heras.
\newblock Icdar 2013 robust reading competition.
\newblock In {\em ICDAR}, pages 1484--1493. IEEE, 2013.

\bibitem{SRResNet}
Christian Ledig, Lucas Theis, Ferenc Huszar, Jose Caballero, Andrew Cunningham,
  Alejandro Acosta, Andrew~P. Aitken, Alykhan Tejani, Johannes Totz, Zehan
  Wang, and Wenzhe Shi.
\newblock Photo-realistic single image super-resolution using a generative
  adversarial network.
\newblock In {\em CVPR}, pages 105--114, 2017.

\bibitem{SATRN:lee2020recognizing}
Junyeop Lee, Sungrae Park, Jeonghun Baek, Seong~Joon Oh, Seonghyeon Kim, and
  Hwalsuk Lee.
\newblock On recognizing texts of arbitrary shapes with 2d self-attention.
\newblock In {\em CVPR Workshops}, pages 546--547, 2020.

\bibitem{SAR:li2019show}
Hui Li, Peng Wang, Chunhua Shen, and Guyu Zhang.
\newblock Show, attend and read: A simple and strong baseline for irregular
  text recognition.
\newblock In {\em AAAI}, pages 8610--8617, 2019.

\bibitem{liao2019scene}
Minghui Liao, Jian Zhang, Zhaoyi Wan, Fengming Xie, Jiajun Liang, Pengyuan Lyu,
  Cong Yao, and Xiang Bai.
\newblock Scene text recognition from two-dimensional perspective.
\newblock In {\em AAAI}, pages 8714--8721, 2019.

\bibitem{SCATTER:litman2020scatter}
Ron Litman, Oron Anschel, Shahar Tsiper, Roee Litman, Shai Mazor, and R
  Manmatha.
\newblock Scatter: selective context attentional scene text recognizer.
\newblock In {\em CVPR}, pages 11962--11972, 2020.

\bibitem{Persec}
Hao Liu, Bin Wang, Zhimin Bao, Mobai Xue, Sheng Kang, Deqiang Jiang, Yinsong
  Liu, and Bo Ren.
\newblock Perceiving stroke-semantic context: Hierarchical contrastive learning
  for robust scene text recognition.
\newblock In {\em AAAI}, pages 1702--1710, 2022.

\bibitem{AdamW}
Ilya Loshchilov and Frank Hutter.
\newblock Decoupled weight decay regularization.
\newblock In {\em ICLR}. OpenReview.net, 2019.

\bibitem{SimAN}
Canjie Luo, Lianwen Jin, and Jingdong Chen.
\newblock Siman: exploring self-supervised representation learning of scene
  text via similarity-aware normalization.
\newblock In {\em CVPR}, pages 1039--1048, 2022.

\bibitem{Masktextspotter}
Pengyuan Lyu, Minghui Liao, Cong Yao, Wenhao Wu, and Xiang Bai.
\newblock Mask textspotter: An end-to-end trainable neural network for spotting
  text with arbitrary shapes.
\newblock In {\em ECCV}, pages 67--83, 2018.

\bibitem{Logo}
Moushumi Medhi, Shubham Sinha, and Rajiv~Ranjan Sahay.
\newblock A text recognition augmented deep learning approach for logo
  identification.
\newblock In {\em Computer Vision, Graphics, and Image Processing - {ICVGIP}},
  volume 10481, pages 145--156, 2016.

\bibitem{IIIT5K:mishra2012scene}
Anand Mishra, Karteek Alahari, and CV Jawahar.
\newblock Scene text recognition using higher order language priors.
\newblock In {\em BMVC}, pages 1--11, 2012.

\bibitem{HAN}
Ben Niu, Weilei Wen, Wenqi Ren, Xiangde Zhang, Lianping Yang, Shuzhen Wang,
  Kaihao Zhang, Xiaochun Cao, and Haifeng Shen.
\newblock Single image super-resolution via a holistic attention network.
\newblock In {\em ECCV}, volume 12357, pages 191--207, 2020.

\bibitem{Latex}
Shuai Peng, Liangcai Gao, Ke Yuan, and Zhi Tang.
\newblock Image to latex with graph neural network for mathematical formula
  recognition.
\newblock In {\em ICDAR}, volume 12822, pages 648--663, 2021.

\bibitem{SP:phan2013recognizing}
Trung~Quy Phan, Palaiahnakote Shivakumara, Shangxuan Tian, and Chew~Lim Tan.
\newblock Recognizing text with perspective distortion in natural scenes.
\newblock In {\em ICCV}, pages 569--576, 2013.

\bibitem{PIMNet}
Zhi Qiao, Yu Zhou, Jin Wei, Wei Wang, Yuan Zhang, Ning Jiang, Hongbin Wang, and
  Weiping Wang.
\newblock Pimnet: a parallel, iterative and mimicking network for scene text
  recognition.
\newblock In {\em ACM MM}, pages 2046--2055, 2021.

\bibitem{CT:risnumawan2014robust}
Anhar Risnumawan, Palaiahankote Shivakumara, Chee~Seng Chan, and Chew~Lim Tan.
\newblock A robust arbitrary text detection system for natural scene images.
\newblock {\em Expert Systems with Applications}, 41(18):8027--8048, 2014.

\bibitem{CRNN:shi2016end}
Baoguang Shi, Xiang Bai, and Cong Yao.
\newblock An end-to-end trainable neural network for image-based sequence
  recognition and its application to scene text recognition.
\newblock {\em IEEE TPAMI}, 39(11):2298--2304, 2016.

\bibitem{TextVQA}
Amanpreet Singh, Vivek Natarjan, Meet Shah, Yu Jiang, Xinlei Chen, Devi Parikh,
  and Marcus Rohrbach.
\newblock Towards vqa models that can read.
\newblock In {\em CVPR}, pages 8317--8326, 2019.

\bibitem{TextOCR}
Amanpreet Singh, Guan Pang, Mandy Toh, Jing Huang, Wojciech Galuba, and Tal
  Hassner.
\newblock Textocr: Towards large-scale end-to-end reasoning for
  arbitrary-shaped scene text.
\newblock In {\em CVPR}, pages 8802--8812, 2021.

\bibitem{wan2020textscanner}
Zhaoyi Wan, Minghang He, Haoran Chen, Xiang Bai, and Cong Yao.
\newblock Textscanner: Reading characters in order for robust scene text
  recognition.
\newblock In {\em AAAI}, pages 12120--12127, 2020.

\bibitem{SVT:wang2011end}
Kai Wang, Boris Babenko, and Serge Belongie.
\newblock End-to-end scene text recognition.
\newblock In {\em ICCV}, pages 1457--1464. IEEE, 2011.

\bibitem{MGP}
Peng Wang, Cheng Da, and Cong Yao.
\newblock Multi-granularity prediction for scene text recognition.
\newblock In {\em ECCV}, volume 13688, pages 339--355, 2022.

\bibitem{DAN:wang2020decoupled}
Tianwei Wang, Yuanzhi Zhu, Lianwen Jin, Canjie Luo, Xiaoxue Chen, Yaqiang Wu,
  Qianying Wang, and Mingxiang Cai.
\newblock Decoupled attention network for text recognition.
\newblock In {\em AAAI}, pages 12216--12224, 2020.

\bibitem{TSRN}
Wenjia Wang, Enze Xie, Xuebo Liu, Wenhai Wang, Ding Liang, Chunhua Shen, and
  Xiang Bai.
\newblock Scene text image super-resolution in the wild.
\newblock In {\em ECCV}, volume 12355, pages 650--666, 2020.

\bibitem{visionLAN}
Yuxin Wang, Hongtao Xie, Shancheng Fang, Jing Wang, Shenggao Zhu, and Yongdong
  Zhang.
\newblock From two to one: A new scene text recognizer with visual language
  modeling network.
\newblock In {\em ICCV}, pages 14194--14203, 2021.

\bibitem{superresolution}
Zhou Wang, A.C. Bovik, H.R. Sheikh, and E.P. Simoncelli.
\newblock Image quality assessment: from error visibility to structural
  similarity.
\newblock {\em IEEE TIP}, 13(4):600--612, 2004.

\bibitem{multimodal}
Jiajia Wu, Jun Du, Fengren Wang, Chen Yang, Xinzhe Jiang, Jinshui Hu, Bing Yin,
  Jianshu Zhang, and Lirong Dai.
\newblock A multimodal attention fusion network with a dynamic vocabulary for
  textvqa.
\newblock {\em PR}, 122:108214, 2022.

\bibitem{xie2022understanding}
Xudong Xie, Ling Fu, Zhifei Zhang, Zhaowen Wang, and Xiang Bai.
\newblock Toward understanding wordart: Corner-guided transformer for scene
  text recognition, 2022.

\bibitem{SimMIM}
Zhenda Xie, Zheng Zhang, Yue Cao, Yutong Lin, Jianmin Bao, Zhuliang Yao, Qi
  Dai, and Han Hu.
\newblock Simmim: a simple framework for masked image modeling.
\newblock In {\em CVPR}, pages 9643--9653, 2022.

\bibitem{TextSeg}
Xingqian Xu, Zhifei Zhang, Zhaowen Wang, Brian Price, Zhonghao Wang, and
  Humphrey Shi.
\newblock Rethinking text segmentation: {A} novel dataset and a text-specific
  refinement approach.
\newblock In {\em CVPR}, pages 12045--12055, 2021.

\bibitem{yan2021primitive}
Ruijie Yan, Liangrui Peng, Shanyu Xiao, and Gang Yao.
\newblock Primitive representation learning for scene text recognition.
\newblock In {\em CVPR}, pages 284--293, 2021.

\bibitem{ScRN:yang2019symmetry}
Mingkun Yang, Yushuo Guan, Minghui Liao, Xin He, Kaigui Bian, Song Bai, Cong
  Yao, and Xiang Bai.
\newblock Symmetry-constrained rectification network for scene text
  recognition.
\newblock In {\em ICCV}, pages 9147--9156, 2019.

\bibitem{DiG}
Mingkun Yang, Minghui Liao, Pu Lu, Jing Wang, Shenggao Zhu, Hualin Luo, Qi
  Tian, and Xiang Bai.
\newblock Reading and writing: Discriminative and generative modeling for
  self-supervised text recognition.
\newblock In {\em ACM MM}, pages 4214--4223, 2022.

\bibitem{TAP}
Zhengyuan Yang, Yijuan Lu, Jianfeng Wang, Xi Yin, Dinei Flor{\^{e}}ncio, Lijuan
  Wang, Cha Zhang, Lei Zhang, and Jiebo Luo.
\newblock {TAP:} text-aware pre-training for text-vqa and text-caption.
\newblock In {\em CVPR}, pages 8751--8761, 2021.

\bibitem{SRN:yu2020towards}
Deli Yu, Xuan Li, Chengquan Zhang, Tao Liu, Junyu Han, Jingtuo Liu, and Errui
  Ding.
\newblock Towards accurate scene text recognition with semantic reasoning
  networks.
\newblock In {\em CVPR}, pages 12113--12122, 2020.

\bibitem{robustscanner}
Xiaoyu Yue, Zhanghui Kuang, Chenhao Lin, Hongbin Sun, and Wayne Zhang.
\newblock Robustscanner: Dynamically enhancing positional clues for robust text
  recognition.
\newblock In {\em ECCV}, pages 135--151. Springer, 2020.

\bibitem{zhang2023relational}
Jinglei Zhang, Tiancheng Lin, Yi Xu, Kai Chen, and Rui Zhang.
\newblock Relational contrastive learning for scene text recognition, 2023.

\bibitem{ConCLR}
Xinyun Zhang, Binwu Zhu, Xufeng Yao, Qi Sun, Ruiyu Li, and Bei Yu.
\newblock Context-based contrastive learning for scene text recognition.
\newblock In {\em AAAI}, pages 3353--3361, 2022.

\bibitem{PCAN}
Cairong Zhao, Shuyang Feng, Brian~Nlong Zhao, Zhijun Ding, Jun Wu, Fumin Shen,
  and Heng~Tao Shen.
\newblock Scene text image super-resolution via parallelly contextual attention
  network.
\newblock In {\em ACM MM}, pages 2908--2917, 2021.

\bibitem{zhong2022sgbanet}
Dajian Zhong, Shujing Lyu, Palaiahnakote Shivakumara, Bing Yin, Jiajia Wu,
  Umapada Pal, and Yue Lu.
\newblock Sgbanet: Semantic gan and balanced attention network for arbitrarily
  oriented scene text recognition.
\newblock In {\em ECCV}, 2022.

\bibitem{IBoT}
Jinghao Zhou, Chen Wei, Huiyu Wang, Wei Shen, Cihang Xie, Alan~L. Yuille, and
  Tao Kong.
\newblock Image {BERT} pre-training with online tokenizer.
\newblock In {\em ICLR}, 2022.

\end{thebibliography}
}

\end{document}